%% file: main.tex
\documentclass[runningheads]{llncs}
\usepackage{graphicx}
\usepackage{comment}
\usepackage{amsmath,amssymb} 
\usepackage{color}

\usepackage{animate}
\usepackage{subfig}
\usepackage{array}
\usepackage{multirow}
\usepackage{mathabx}
\usepackage[hidelinks]{hyperref}
\usepackage{enumitem}
\usepackage{xspace}
\usepackage{xcolor}
\usepackage{color,soul}
\usepackage{multicol}
\usepackage{colortbl}
\usepackage{tabularx,booktabs}
\usepackage{cancel}
\usepackage{booktabs}
\usepackage{chngcntr}
\newcommand{\sota}{state-of-the-art\xspace}
\newcommand{\vidtovid}{vid2vid\xspace}
\newcommand{\flow}{Ours w/o W.C.}
\newcommand{\guidance}{Ours}
\newcommand{\rotangle}{90}

\newcommand{\website}{https://nvlabs.github.io/wc-vid2vid/arxiv_videos/}

\newcommand{\x}{\tilde{\mathbf{x}}}

\newcommand{\g}{\tilde{\mathbf{g}}}
\newcommand{\bx}{{\mathbf{x}}}
\newcommand{\bs}{{\mathbf{s}}}
\newcommand{\bw}{{\mathbf{w}}}

\makeatletter
\DeclareRobustCommand\onedot{\futurelet\@let@token\@onedot}
\def\@onedot{\ifx\@let@token.\else.\null\fi\xspace}
\def\eg{\emph{e.g}\onedot} 
\def\ie{\emph{i.e}\onedot} 
 
\def\etc{\emph{etc}\onedot} 
 
\def\etal{\emph{et al}\onedot}
\makeatother

\let\oldFootnote\footnote
\newcommand\nextToken\relax
\renewcommand\footnote[1]{%
    \oldFootnote{#1}\futurelet\nextToken\isFootnote}
\newcommand\isFootnote{%
    \ifx\footnote\nextToken\textsuperscript{,}\fi}
\newcommand\blfootnote[1]{%
  \begingroup
  \renewcommand\thefootnote{}\footnote{#1}%
  \addtocounter{footnote}{-1}%
  \endgroup
}

\definecolor{overview_orange}{HTML}{FFA726}
\definecolor{overview_green}{HTML}{9CCC65}
\definecolor{overview_purple}{HTML}{D877E9}
\definecolor{overview_red}{HTML}{EF5350}
\definecolor{overview_gray}{HTML}{BEBEBE}

\newcolumntype{Y}{>{\centering\arraybackslash}X}

\begin{document}
\pagestyle{headings}
\mainmatter
\def\ECCVSubNumber{587}  

\title{World-Consistent Video-to-Video Synthesis}
\titlerunning{World-Consistent Video-to-Video Synthesis}
\author{Arun Mallya$^\ast$ \and
Ting-Chun Wang$^\ast$ \and
Karan Sapra \and
Ming-Yu Liu}
\authorrunning{A. Mallya et al.}
\institute{NVIDIA \\
\email{\{amallya,tingchunw,ksapra,mingyul\}@nvidia.com} \\
\color{blue}{\url{https://nvlabs.github.io/wc-vid2vid/}}
\blfootnote{$^\ast$ Equal contribution}
}

\maketitle

\input{figures/mannequin_teaser.tex}
\input{src/abstract.tex}

\input{src/introduction.tex}
\input{src/related_work.tex}
\input{src/method.tex}

\input{src/experiments.tex}

\input{src/applications.tex}
\input{src/conclusions.tex}

\clearpage


%
%
\bibliographystyle{splncs04}
\bibliography{guigan}
\clearpage

\appendix
\input{objectives}
\input{network}

\input{additional_results}

\end{document}

%% file: figures/mannequin_teaser.tex

\begin{table}[ht!]
    \setlength{\tabcolsep}{0pt}
    \setlength{\belowcaptionskip}{2pt}
    \centering

    \resizebox{\columnwidth}{!}{%
    \begin{tabular}{c}
        \parbox{\textwidth}{\centering \cellcolor{green!50} \bf Illustration of an input 3D source world} \\
        \includegraphics[width=0.8\textwidth, trim={0, 1cm, 0, 0}, clip]{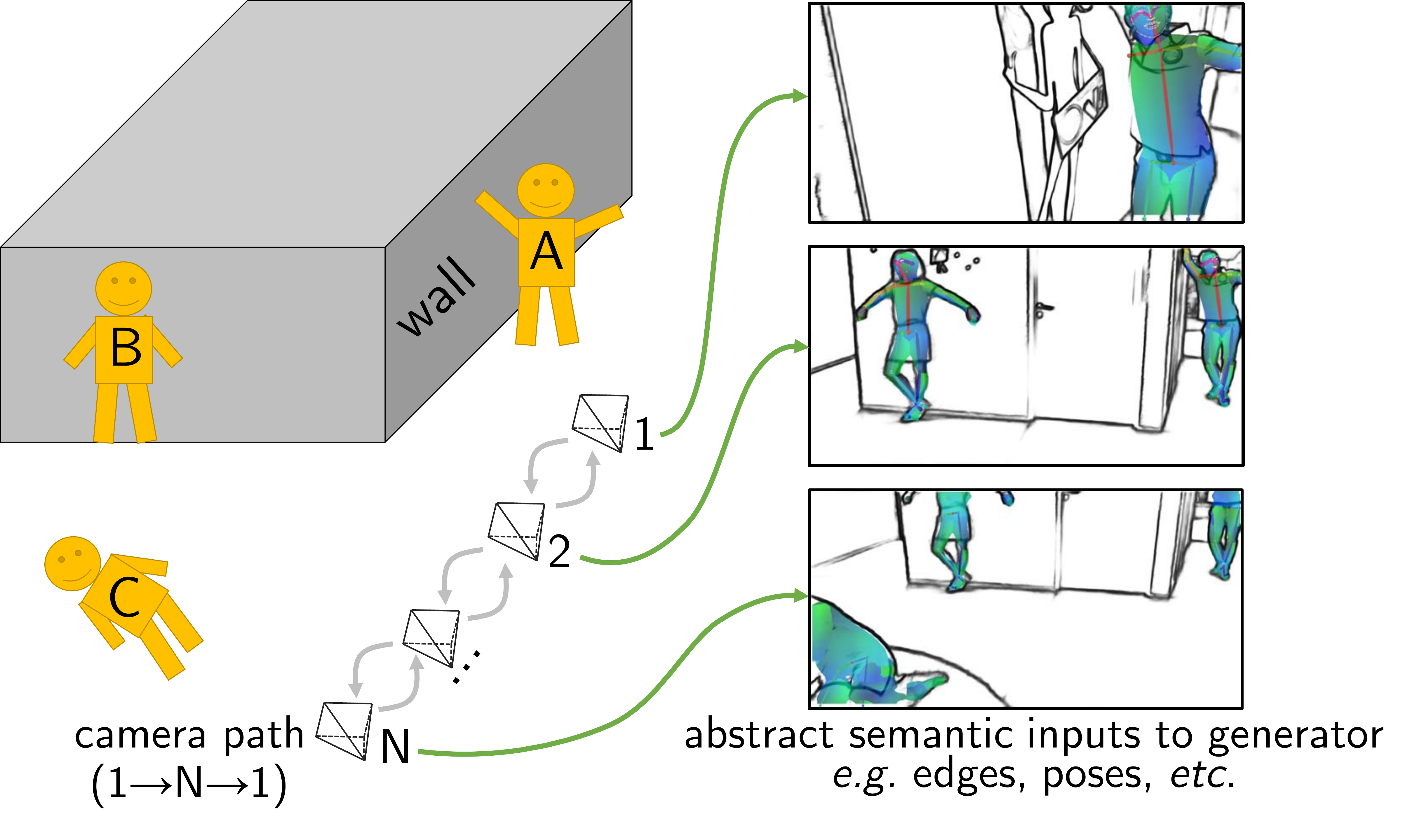}
    \end{tabular}
    }

    \resizebox{\columnwidth}{!}{%
    \begin{tabular}{c@{\hskip 2pt}ccc}
        \multicolumn{4}{c}{\cellcolor{red!50} \bf Generated video outputs based on above inputs} \\
        & \parbox{0.33\textwidth}{\centering First location} & \parbox{0.33\textwidth}{\centering Round-trip video $\circlearrowright$} & \parbox{0.33\textwidth}{\centering Re-visited first location} \\

        {\normalsize \rotatebox[origin=c]{\rotangle}{\centering \bf Ours}} &
        \raisebox{-.5\height}{\includegraphics[width=0.33\textwidth]{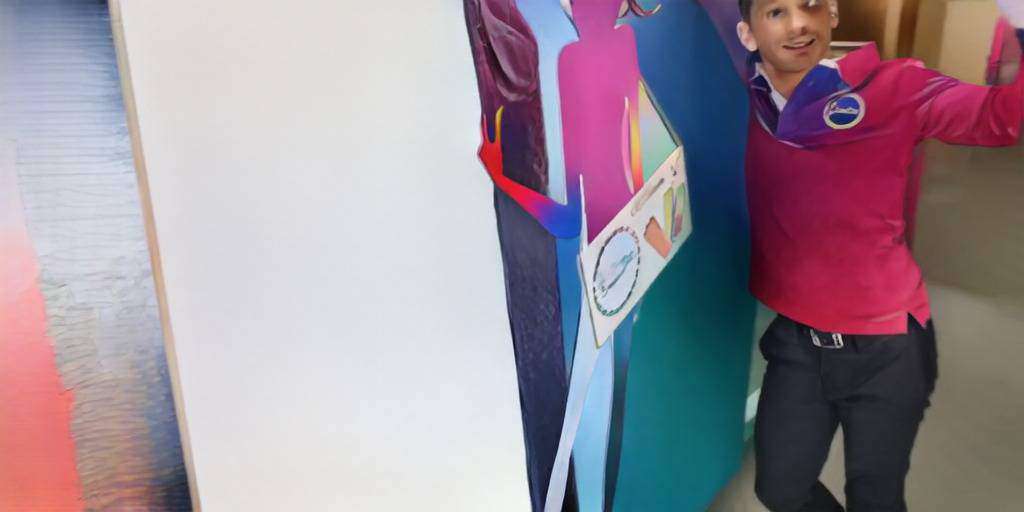}} &
        \multirow[t]{3}{*}{\raisebox{-.833\height}{\href{\website all_stack.mp4}{\includegraphics[width=0.33\textwidth]{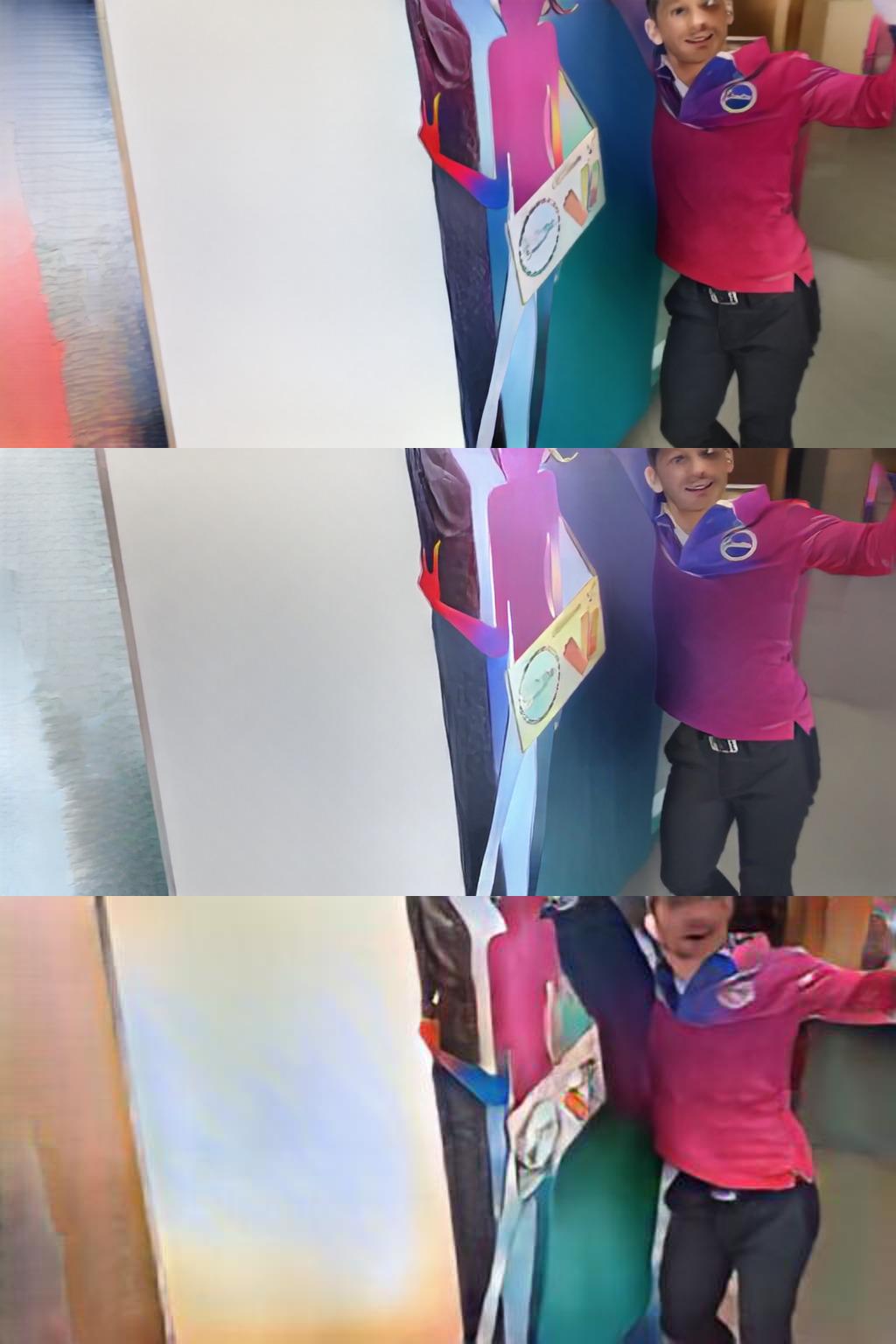}}}} &
        \raisebox{-.5\height}{\includegraphics[width=0.33\textwidth]{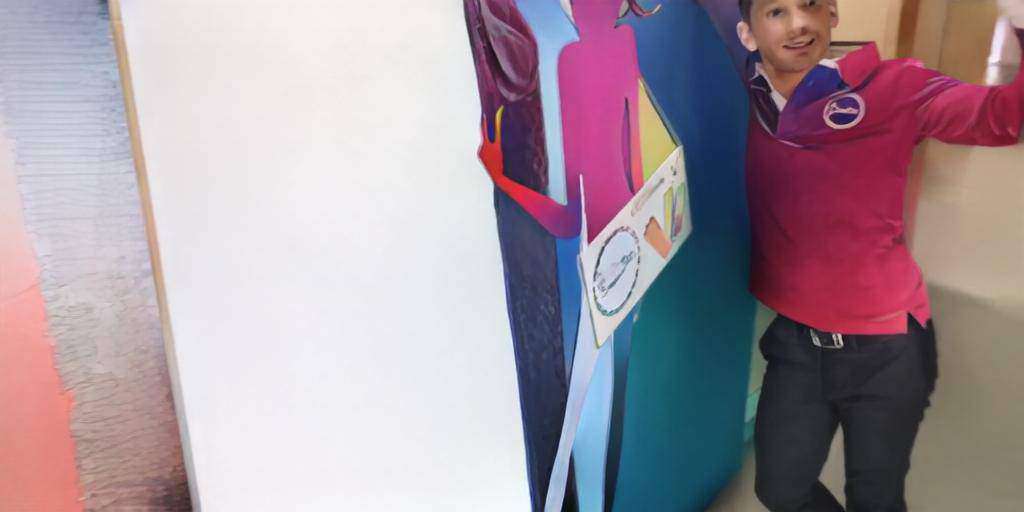}}
         \\

        {\small \rotatebox[origin=c]{\rotangle}{\centering
            \begin{tabular}{@{}c@{}}
                Ours w/o \\
                {\scriptsize World Cons.}
            \end{tabular}
        }} &
        \raisebox{-.5\height}{\includegraphics[width=0.33\textwidth]{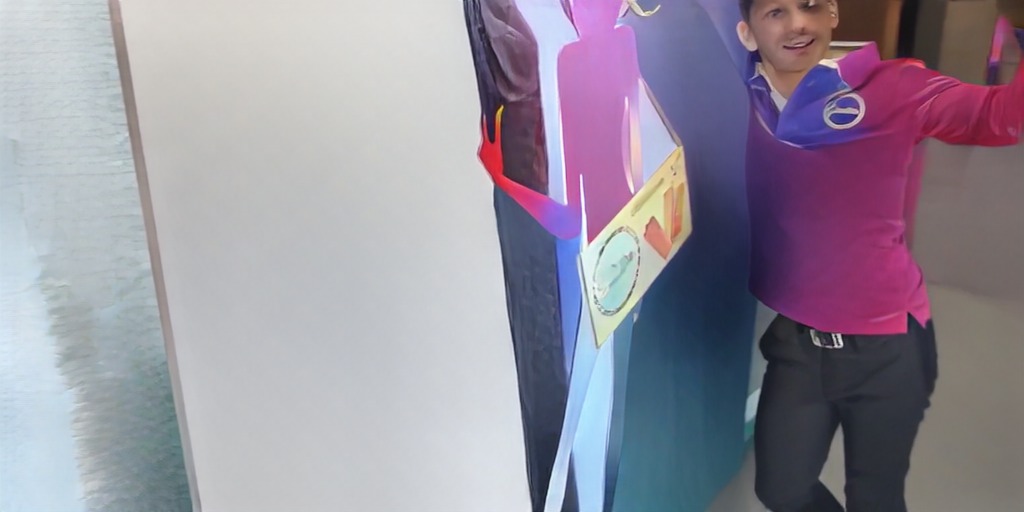}} & &
        \raisebox{-.5\height}{\includegraphics[width=0.33\textwidth]{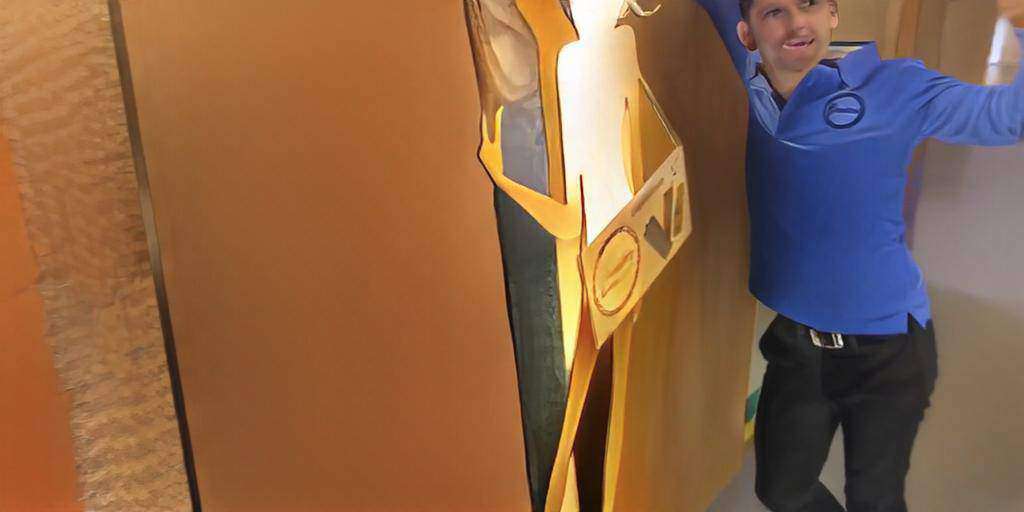}} \\

        {\small \rotatebox[origin=c]{\rotangle}{\centering
            \begin{tabular}{@{}c@{}}
                \vidtovid \\
                \cite{wang2018video}
            \end{tabular}
        }} &
        \raisebox{-.5\height}{\includegraphics[width=0.33\textwidth]{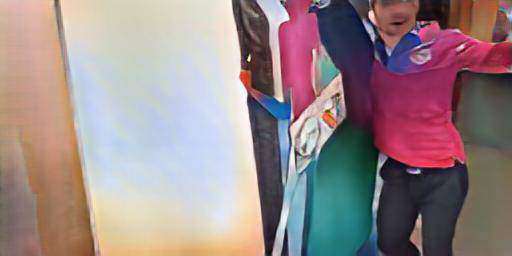}} & &
        \raisebox{-.5\height}{\includegraphics[width=0.33\textwidth]{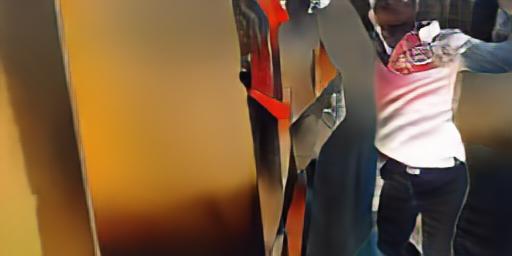}} \\
    \\
    \end{tabular}
    }

    \captionof{figure}{Imagine moving around in a world such as the one abstracted at the top. As you move from locations 1$\rightarrow$N$\rightarrow$1, you would expect the appearance of previously seen walls and people to remain unchanged.
    However, current video-to-video synthesis methods such as vid2vid~\cite{wang2018video} or our improved architecture combining vid2vid with SPADE~\cite{park2019semantic} cannot produce such world-consistent videos (third and second rows).
    Only our method is able to produce videos consistent over viewpoints by adding a mechanism for world consistency (first row).}

    \emph{\normalsize Click any middle column image to play video.}
    \label{fig:teaser}
\end{table}

%% file: src/abstract.tex

\begin{abstract}
Video-to-video synthesis (\vidtovid) aims for converting high-level semantic inputs to photorealistic videos. While existing \vidtovid methods can achieve short-term temporal consistency, they fail to ensure the long-term one. This is because they lack knowledge of the 3D world being rendered and generate each frame only based on the past few frames. To address the limitation, we introduce a novel \vidtovid framework that efficiently and effectively utilizes all past generated frames during rendering. This is achieved by \emph{condensing} the 3D world rendered so far into a physically-grounded estimate of the current frame, which we call the \emph{guidance image}. We further propose a novel neural network architecture to take advantage of the information stored in the guidance images. Extensive experimental results on several challenging datasets verify the effectiveness of our approach in achieving \emph{world consistency}---the output video is consistent within the entire rendered 3D world.
\keywords{neural rendering, video synthesis, GAN}
\end{abstract}

%% file: src/introduction.tex

\section{Introduction}

Video-to-video synthesis~\cite{wang2018video} concerns generating a sequence of photorealistic images given a sequence of semantic representations extracted from a source 3D world. For example, the representations can be the semantic segmentation masks rendered by a graphics engine while driving a car in a virtual city~\cite{wang2018video}. The representations can also be the pose maps extracted from a source video of a person dancing, and the application is to create a video of a different person performing the same dance~\cite{chan2019everybody}. From the creation of a new class of digital artworks to applications in computer graphics, the video-to-video synthesis task has many exciting practical use-cases. A key requirement of any such video-to-video synthesis model is the ability to generate images that are not only individually photorealistic, but also temporally smooth. Moreover, the generated images have to follow the geometric and semantic structure of the source 3D world.

While we have observed steady improvement in photorealism and short-term temporal stability in the generation results, we argue that one crucial aspect of the problem has been largely overlooked, which is the \emph{long-term temporal consistency} problem. As a specific example, when visiting the same location in the virtual city, an existing \vidtovid method~\cite{wang2019few,wang2018video} could generate an image that is very different from the one it generated when the car first visited the location, despite using the same semantic inputs. Existing \vidtovid methods rely on optical flow warping and generate an image conditioned on the past few generated images. While such operations can ensure short-term temporal stability, they cannot guarantee long-term temporal consistency. Existing \vidtovid models have no knowledge of what they have rendered in the past. Even for a short round-trip in a virtual room, these methods fail to preserve the appearances of the wall and the person in the generated video, as illustrated in Fig.~\ref{fig:teaser}.

In this paper, we attempt to address the long-term temporal consistency problem, by bolstering \vidtovid models with memories of the past frames. By combining ideas from scene flow~\cite{vedula1999three} and conditional image synthesis models~\cite{park2019semantic}, we propose a novel architecture that explicitly enforces consistency in the entire generated sequence. We perform extensive experiments on several benchmark datasets, with comparisons to the \sota methods. Both quantitative and visual results verify that our approach achieves significantly better image quality and long-term temporal stability. On the application side, we also show that our approach can be used to generate videos consistent across multiple viewpoints, enabling simultaneous multi-agent world creation and exploration.

%% file: src/related_work.tex

\section{Related work}

{\bf Semantic Image Synthesis}~\cite{chen2017photographic,liu2019learning,park2019semantic,qi2018semi,wang2018high} refers to the problem of converting a single input semantic representation to an output photorealistic image. Built on top of the generative adversarial networks (GAN)~\cite{goodfellow2014generative} framework, existing  methods~\cite{liu2019learning,park2019semantic,wang2018high} propose various novel network architectures to advance \sota. Our work is built on the SPADE architecture proposed by Park \etal~\cite{park2019semantic} but focuses on the temporal stability issue in video synthesis.

\smallskip
\noindent{\bf Conditional GANs} synthesize data conditioned on user input. This stands in contrast to unconditional GANs that synthesize data solely based on random variable inputs~\cite{goodfellow2014generative,gulrajani2017improved,karras2017progressive,karras2018style}. Based on the input type, there exist label-conditional GANs~\cite{brock2018large,miyato2018cgans,odena2016conditional,zhang2019self}, text-conditional GANs \cite{reed2016generative,xu2018attngan,zhang2017stackgan}, image-conditional GANs \cite{benaim2018one,bousmalis2016unsupervised,choi2017stargan,huang2018multimodal,isola2017image,lee2018diverse,liu2016unsupervised,liu2019few,shrivastava2016learning,taigman2016unsupervised,zhu2017unpaired}, scene-graph conditional GANs \cite{johnson2018image}, and layout-conditional GANs \cite{zhao2019image}. Our method is a video-conditional GAN, where we generate a video conditioned on an input video. We address the long-term temporal stability issue that the \sota overlooks~\cite{chan2019everybody,wang2019few,wang2018video}.

\smallskip
\noindent{\bf Video synthesis} exists in many forms, including 1) unconditional video synthesis~\cite{saito2017temporal,tulyakov2017mocogan,vondrick2016generating}, which converts random variable inputs to video clips, 2) future video prediction~\cite{denton2017unsupervised,finn2016unsupervised,hao2018controllable,hu2018video,kalchbrenner2016video,lee2018stochastic,li2018flow,liang2017dual,lotter2016deep,mathieu2015deep,pan2019video,srivastava2015unsupervised,villegas2017decomposing,walker2016uncertain,walker2017pose,xue2016visual}, which generates future video frames based on the observed ones, and 3) video-to-video synthesis~\cite{chan2019everybody,chen2019mocycle,gafni2019vid2game,wang2019few,wang2018video,zhou2019dance}, which converts an input semantic video to a real video. Our work belongs to the last category. Our method treats the input video as one from a self-consistent world so that when the agent returns to a spot that it has previously visited, the newly generated frames should be consistent with the past generated frames. While a few works have focused on improving the temporal consistency of an input video~\cite{bonneel2015blind,lai2018learning,yao2017occlusion}, our method does not treat consistency as a post-processing step, but rather as a core part of the video generation process.

\smallskip
\noindent{\bf Novel-view synthesis} aims to synthesize images at unseen viewpoints given some viewpoints of the scene. Most of the existing works require images at multiple reference viewpoints as input~\cite{choi2019extreme,flynn2019deepview,flynn2016deepstereo,hedman2018deep,kalantari2016learning,mildenhall2019local,zhou2018stereo}. While some works can synthesize novel views based on a single image~\cite{srinivasan2017learning,wiles2019synsin,xie2016deep3d}, the synthesized views are usually close to the reference views. Our work differs from these works in the sense that our input is different -- instead of using a set of RGB images, our network takes in a sequence of semantic maps. If we directly treat all past synthesized frames as reference views, it makes the memory requirement grow linearly with respect to the video length. If we only use the latest frames, the system cannot handle long-term consistency as shown in Fig.~\ref{fig:teaser}. Instead, we propose a novel framework to keep track of the synthesis history in this work.

The closest related works are those on neural rendering~\cite{aliev2019neural,meshry2019neural,sitzmann2019deepvoxels,thies2019deferred}, which can re-render a scene from arbitrary viewpoints after training on a set of given viewpoints. However, note that these methods still require RGB images from different viewpoints as input, making it unsuitable for applications such as those to game engines. On the other hand, our method can directly generate RGB images using semantic inputs, so rendering a virtual world becomes more effortless. Moreover, they need to train a separate model (or part of the model) for each scene, while we only need one model per dataset, or domain.

%% file: src/method.tex

\section{World-consistent video-to-video synthesis}
\label{sec:method}

\input{src/method/background}
\input{src/method/guidance_map_generation.tex}
\input{src/method/framework.tex}

%% file: src/method/background.tex
\medskip
\noindent{\bf Background.} Recent image-to-image translation methods perform extremely well when turning semantic images to realistic outputs. To produce videos instead of images, simply doing it frame-by-frame will usually result in severe flickering artifacts~\cite{wang2018video}. To resolve this, \vidtovid~\cite{wang2018video} proposes to take both the semantic inputs and $L$ previously generated frames as input to the network (\eg $L=3$). The network then generates three outputs -- a hallucinated frame, a flow map, and a (soft) mask. The flow map is used to warp the previous frame and linearly combined with the hallucinated frame using the soft mask. Ideally, the network should reuse the content in the warped frame as much as possible, and only use the disoccluded parts from the hallucinated frame.

While the above framework reduces flickering between neighboring frames, it still struggles to ensure long-term consistency. This is because it only keeps track of the past $L$ frames, and cannot memorize everything in the past. Consider the scenario in Fig.~\ref{fig:teaser}, where an object moves out of and back in the field-of-view. In this case, we would want to make sure its appearance is similar during the revisit, but that cannot be handled by existing frameworks like \vidtovid~\cite{wang2018video}.

In light of this, we propose a new framework to handle \emph{world-consistency}. It is a superset of \emph{temporal consistency}, which only ensures consistency between frames in a video. A world-consistent video should not only be temporally stable, but also be consistent across the entire 3D world the user is viewing. This not only makes the output look more realistic, but also enables applications such as the multi-player scenario where different players can view the same scene from different viewpoints. We achieve this by using a novel \emph{guidance image} conditional scheme, which is detailed below.

%% file: src/method/guidance_map_generation.tex


\input{figures/guidance_generation.tex}
\medskip
\noindent{\bf Guidance images and their generation.} The lack of knowledge about the world structure being generated limits the ability of \vidtovid to generate view-consistent outputs. As shown in Fig.~\ref{fig:cityscapes_results} and Sec.~\ref{sec:experiments}, the color and structure of the objects generated by vid2vid~\cite{wang2018video} tend to drift over time. We believe that in order to produce realistic outputs that are consistent over time and viewpoint change, an ideal method must be aware of the 3D structure of the world.

To achieve this, we introduce the concept of ``\emph{guidance images}'', which are physically-grounded estimates of what the next output frame should look like, based on how the world has been generated so far. As alluded to in their name, the role of these ``\emph{guidance images}'' is to guide the generative model to produce colors and textures that respect previous outputs. Prior works including \vidtovid~\cite{wang2018video} rely on optical flows to warp the previous frame for producing an estimate of the next frame. Our guidance image differs from this warped frame in two aspects. First, instead of using optical flow, the guidance image should be generated by using the motion field, or scene flow, which describes the true motion of each 3D point in the world\footnote{As an example, consider a textureless sphere rotating under constant illumination. In this case, the optical flow would be zero, but the motion field would be nonzero.}.
Second, the guidance image should aggregate information from \emph{all} past viewpoints (and thus frames), instead of only the direct previous frames as in \vidtovid. This makes sure that the generated frame is consistent with the entire history.

While estimating motion fields without an RGB-D sensor~\cite{golyanik2017multiframe} or a rendering engine~\cite{dosovitskiy2017carla} is not easy, we can obtain motion fields for the static parts of the world by reconstructing part of the 3D world using structure from motion (SfM)~\cite{longuet1981computer,tomasi1992shape}. This enables us to generate guidance images as shown in Fig.~\ref{fig:guidance_generation} for training our video-to-video synthesis method using datasets captured by regular cameras. Once we have the 3D point cloud of the world, the video synthesis process can be thought of as a camera moving through the world and texturing every new 3D point it sees. Consider a camera moving through space and time as shown in the left part of Fig.~\ref{fig:guidance_generation}. Suppose we generate an output image at $t=0$. This image can be back-projected to the 3D point cloud and colors can be assigned to the points, so as to create a persistent representation of the world. At a later time step, $t=N$, we can obtain the projection of the 3D point cloud to the camera and create a guidance image leveraging estimated motion fields. Our method can then generate an output frame based on the guidance image.

Although we generate guidance images using the projection of 3D point clouds, it can also be generated by any other method that gives a reasonable estimate. This makes the concept powerful, as we can use different sources to generate guidance images at training and test time. For example, at test time we can generate guidance images using a graphics engine, which can provide ground truth 3D correspondences. This enables just-in-time colorization of a virtual 3D world with real-world colors and textures, as we move through the world.

Note that our guidance image also differs from the projected image used in prior works like Meshry~\etal~\cite{meshry2019neural} in several aspects. First, in their case, the 3D point cloud is fixed once constructed, while in our case it is constantly being ``colorized" as we synthesize more and more frames. As a result, our guidance image is blank at the beginning, and can become denser depending on the viewpoint.
Second, the way we use these guidance images to generate outputs is also different. The guidance images can have misalignments and holes due to limitations of SfM, for example in the background and in the person's head in Fig.~\ref{fig:guidance_generation}. As a result, our method also differs from DeepFovea~\cite{kaplanyan2019deepfovea}, which inpaints sparsely but accurately rendered video frames.
In the following subsection, we describe a method that is robust to noises in guidance images, so it can produce outputs consistent over time and viewpoints.

%% file: figures/guidance_generation.tex
\begin{figure}[t!]
    \centering
    \includegraphics[width=0.95\textwidth]{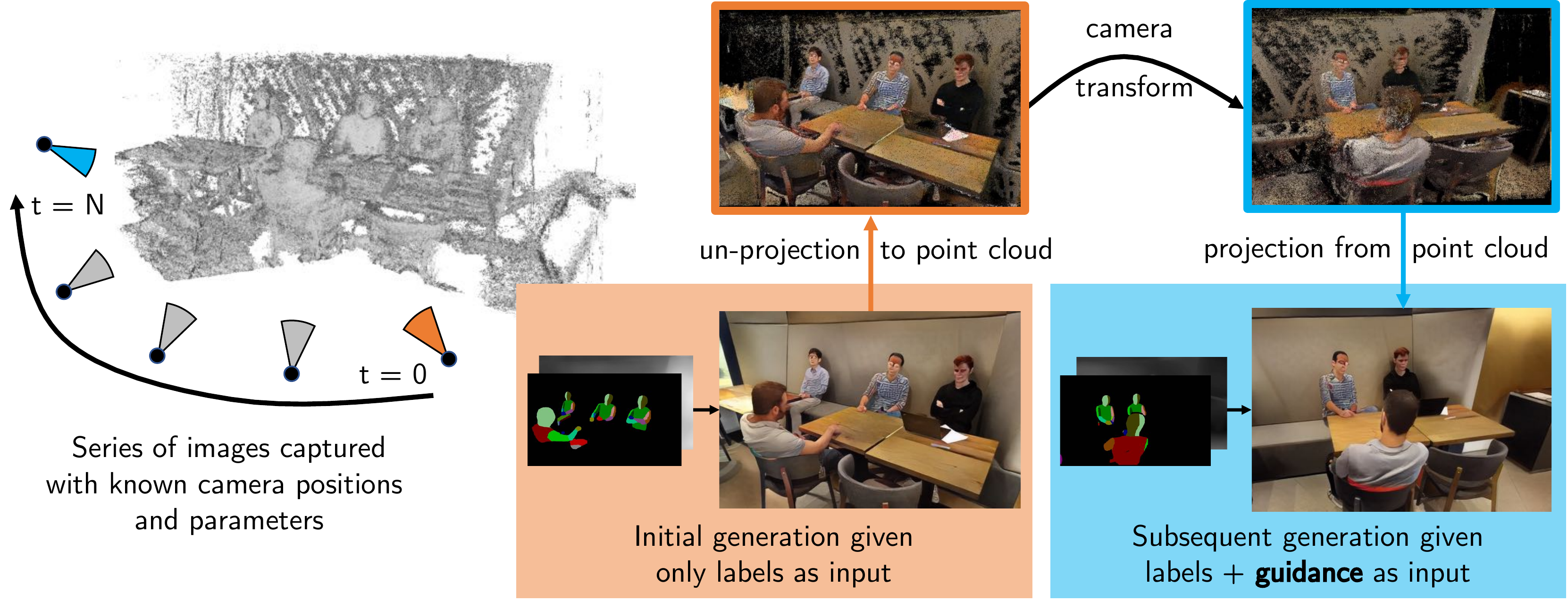}
    \caption{Overview of \emph{guidance image} generation for training.
    	Consider a scene in which a camera(s) with known parameters and positions travels over time $t=0, \cdots, N$. At $t=0$, the scene is textureless and an output image is generated for this viewpoint. The output image is then back-projected to the scene and a \emph{guidance image} for a subsequent camera position is generated by projecting the partially textured point cloud. Using this guidance image, the generative method can produce an output that is consistent across views and smooth over time. Note that the guidance image can be noisy, misaligned, and have holes, and the generation method should be robust to such inputs.
    }
    \label{fig:guidance_generation}
\end{figure}

%% file: src/method/framework.tex


\input{figures/method_overview.tex}
\medskip
\noindent{\bf Framework for generating videos using guidance images.} Once the guidance images are generated, we are able to utilize them to synthesize the next frame. Our generator network is based on the SPADE architecture proposed by Park \etal~\cite{park2019semantic}, which accepts a random vector encoding the image style as input and uses a series of SPADE blocks and upsampling layers to generate an output image. Each SPADE block takes a semantic map as input and learns to modulate the incoming feature maps through an affine transform $y = x \cdot \gamma_\textrm{seg} + \beta_\textrm{seg}$, where $x$ is the incoming feature map, and $\gamma_\textrm{seg}$ and $\beta_\textrm{seg}$ are predicted from the input segmentation map.

An overview of our method is shown in Fig.~\ref{fig:method_overview}.
At a high-level, our method consists of four sub-networks: 1) an input label embedding network (\textcolor{overview_orange}{\bf orange}), 2) an image encoder (\textcolor{overview_red}{\bf red}), 3) a flow embedding network (\textcolor{overview_green}{\bf green}), and 4) an image generator (\textcolor{overview_gray}{\bf gray}). In our method, we make two modifications to the original SPADE network. First, we feed in the concatenated labels (semantic segmentation, edge maps, \etc) to a label embedding network (\textcolor{overview_orange}{\bf orange}), and extract features in corresponding output layers as input to each SPADE block in the generator. Second, to keep the image style consistent over time, we encode the previously synthesized frame using the image encoder (\textcolor{overview_red}{\bf red}), and provide this embedding to our generator (\textcolor{overview_gray}{\bf gray}) in place of the random vector\footnote{When generating the first frame where no previous frame exists, we use an encoder which accepts the semantic map as input.}.

\medskip\noindent{\it Utilizing guidance images.}
Although using this modified SPADE architecture produces output images with better visual quality than \vidtovid~\cite{wang2018video}, the outputs are not temporally stable, as shown in Sec.~\ref{sec:experiments}. To ensure world-consistency of the output, we would want to incorporate information from the introduced guidance images. Simply linearly combining it with the hallucinated frame from the SPADE generator is problematic, since the hallucinated frame may contain something very different from the guidance images. Another way is to directly concatenate it with the input labels. However, the semantic inputs and guidance images have different physical meanings. Besides, unlike semantic inputs, which are labeled densely (per pixel), the guidance images are labeled sparsely. Directly concatenating them would require the network to compensate for the difference. Hence, to avoid these potential issues, we choose to treat these two types of inputs differently.

To handle the sparsity of the guidance images, we first apply partial convolutions~\cite{liu2018image} on these images to extract features. Partial convolutions only convolve valid regions in the input with the convolution kernels, so the output features can be uncontaminated by the holes in the image. These features are then used to generate affine transformation parameters $\gamma_\textrm{guidance}$ and $\beta_\textrm{guidance}$, which are \emph{inserted} into existing SPADE blocks while keeping the rest of the blocks untouched. This results in a \emph{Multi-SPADE} module, which allows us to use multiple conditioned inputs in sequence, so we can not only condition on the current input labels, but also on our guidance images,
\begin{equation}
\begin{aligned}
y   &= (x    \cdot \gamma_\textrm{label} + \beta_\textrm{label})  \cdot \gamma_\textrm{guidance} + \beta_\textrm{guidance}.
\end{aligned}
\label{eq:multi_spade}
\end{equation}

Using this module yields several benefits. First, conditioning on these maps generates more temporally smooth and higher quality frames than simple linear blending techniques. Separating the two types of input (semantic labels and guidance images) also allows us to adopt different types of convolutions (i.e.\ normal vs.\ partial). Second, since most of the network architecture remains unchanged, we can initialize the weights of the generator with one trained for single image generation. It is easy to collect large training datasets for single image generation by crawling the internet, while video datasets can be harder to collect and annotate. After the single image generator is trained, we can train a video generator by just training the newly added layers (\ie\ layers generating $\gamma_\textrm{guidance}$ and $\beta_\textrm{guidance}$) and only finetune the other parts of the network.

\medskip
\noindent{\it Handling dynamic objects.} The guidance image allows us to generate world-consistent outputs over time. However, since the guidance is generated based on SfM for real-world scenes, it has the inherent limitation that SfM cannot handle dynamic objects. To resolve this issue, we revert to using optical flow-warped frames to serve as additional maps in addition to the guidance images we have from SfM. The complete Multi-SPADE module then becomes
\begin{equation}
\begin{aligned}
y   &=  \big{(}(x    \cdot \gamma_\textrm{label} + \beta_\textrm{label})  \cdot \gamma_\textrm{flow} + \beta_\textrm{flow}\big{)}  \cdot \gamma_\textrm{guidance} + \beta_\textrm{guidance},
\end{aligned}
\label{eq:multi_spade}
\end{equation}
where $\gamma_\textrm{flow}$ and $\beta_\textrm{flow}$ are generated using a flow-embedding network (\textcolor{overview_green}{\bf green}) applied on the optical flow-warped previous frame. This provides additional constraints that the generated frame should be consistent even in the dynamic regions. Note that this is needed only due to the limitation of SfM, and can potentially be removed when ground truth / high quality 3D registrations are available, for example in the case of game engines, or RGB-D data capture.

\input{figures/input_output.tex}

Figure~\ref{fig:input_output} shows a sample set of inputs and outputs generated by our method on the Cityscapes dataset.

%% file: figures/method_overview.tex
\begin{figure}[t!]
    \centering
    \includegraphics[width=\columnwidth, trim=0 0.75cm 0 0, clip]{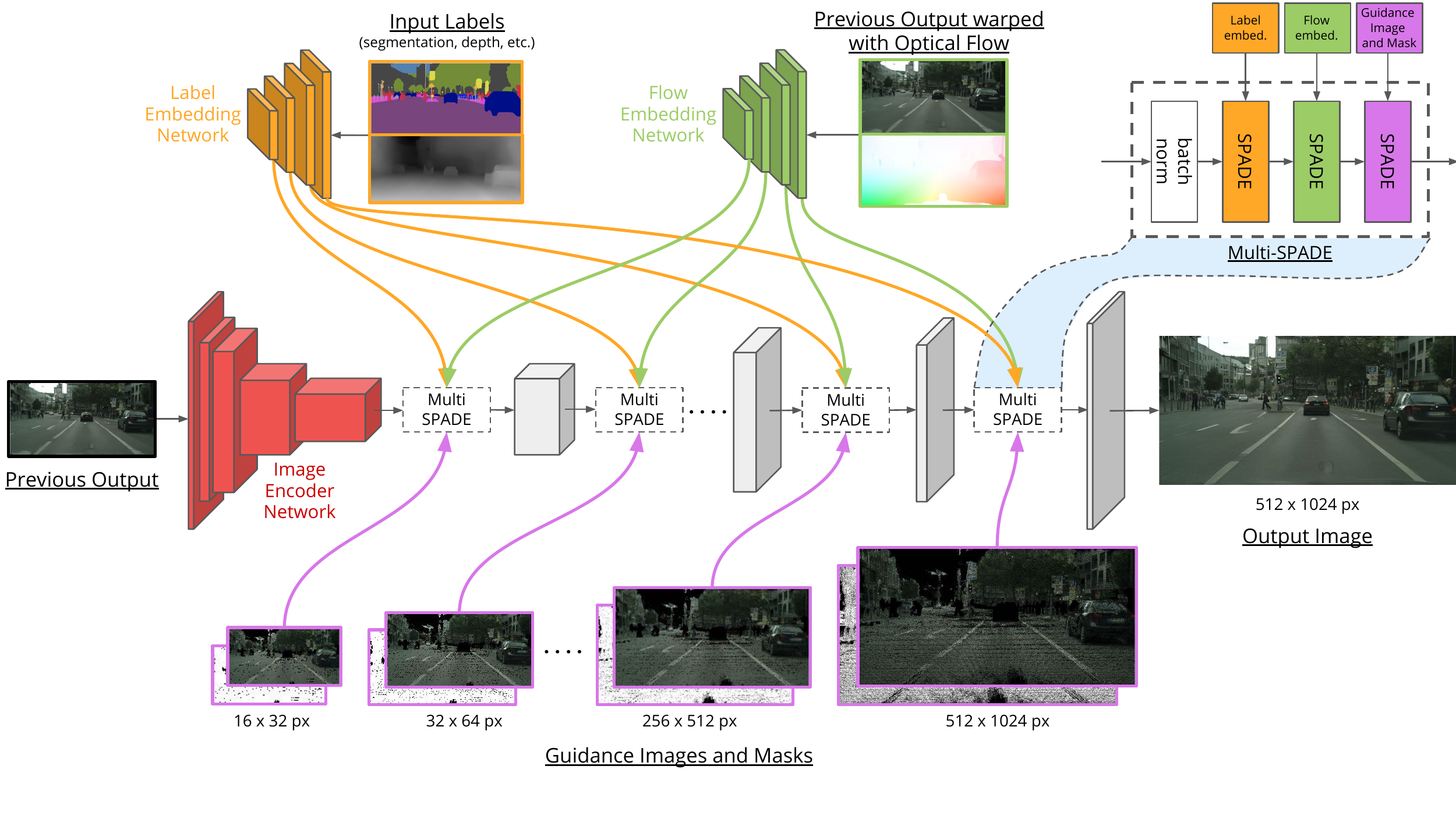}
    \caption{Overview of our world consistent video-to-video synthesis architecture. Our Multi-SPADE module takes input labels, warped previous frames, and guidance images to modulate the features in each layer of our generator.}
    \label{fig:method_overview}
\end{figure}

%% file: figures/input_output.tex

\begin{table}[t!]
    \centering
    \setlength{\tabcolsep}{0pt}
    \begin{tabular}{cccc}
        \parbox{0.25\textwidth}{\centering Segmentation} &
        \parbox{0.25\textwidth}{\centering Depth} &
        \parbox{0.25\textwidth}{\centering Guidance Image} &
        \parbox{0.25\textwidth}{\centering Generated Output} \\
        \multicolumn{4}{c}{\href{\website inputs_outputs.mp4}{\includegraphics[width=\textwidth]{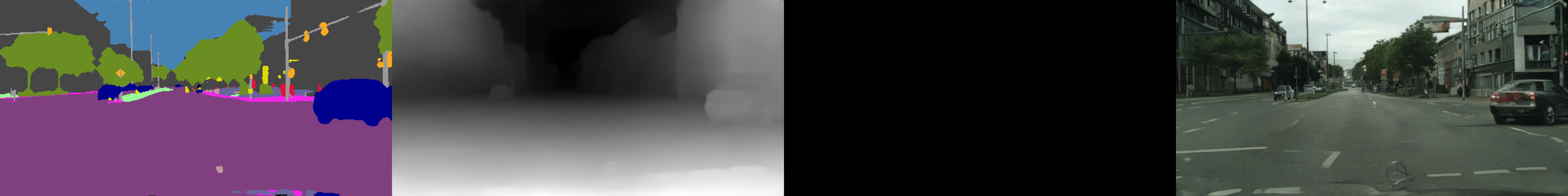}}}
    \end{tabular}
    \captionof{figure}{Sample inputs and generated outputs on Cityscapes. Note how the guidance image is initially black, and becomes denser as more frames are synthesized. Click on any image to play video.}
    \label{fig:input_output}
\end{table}

%% file: src/experiments.tex

\section{Experiments}
\label{sec:experiments}

\noindent\textbf{Implementation details.} We train our network in two stages. In the first stage, we only train our network to generate single images. This means that only the first SPADE layer of our Multi-SPADE block (visualized in Fig.~\ref{fig:method_overview}) is trained. Following this, we have a network that can generate high-quality single frame outputs. In the second stage, we train on video clips, progressively doubling the generated video length every epoch, starting from 8 frames and stopping at 32 frames. In this stage, all 3 SPADE layers of each Multi-SPADE block are trained. We found that this two-stage pipeline makes the training faster and more stable.
We observed that the ordering of the flow and guidance SPADEs did not make a significant difference in the output quality.
We train the network for 20 epochs in each stage, and this takes about 10 days on an NVIDIA DGX-1 (8 V-100 GPUs) for an output resolution of $1024\times 512$.

We train our generator with the multi-scale image discriminator using perceptual and GAN feature matching losses as in SPADE~\cite{park2019semantic}. Following vid2vid~\cite{wang2018video}, we add a temporal video discriminator at two temporal scales and a warping loss that encourages the output frame to be similar to the optical flow-warped previous frame. We also add a loss term to encourage the output frame to correspond to the guidance image, and this is necessary to ensure view consistency. Additional details about architecture and loss terms can be found in Appendix~\ref{appendix:objective_functions} and~\ref{appendix:network}. Code and trained models will be released upon publication.

\medskip

\noindent\textbf{Datasets.} We train and evaluate our method on three datasets, Cityscapes~\cite{Cordts2016cityscapes}, MannequinChallenge~\cite{li2019learning}, and ScanNet~\cite{dai2017scannet}, as they have mostly static scenes where existing SfM methods perform well.
\begin{itemize}[label=\textbullet, topsep=2pt, itemsep=2pt]
    \item \textbf{Cityscapes}~\cite{Cordts2016cityscapes}. This dataset consists of driving videos of $2048\times 1024$ resolution captured in several German cities, using a pair of stereo cameras. We split this dataset into a training set of 3500 videos with 30 frames each, and a test set of 3 long sequences with 600-1200 frames each, similar to \vidtovid~\cite{wang2018video}. As not all the images are labeled with segmentation masks, we annotate the images using the network from Zhu \etal~\cite{zhu2019improving}, which is based on a DeepLabv3-Plus~\cite{chen2018encoder}-like architecture with a WideResNet38~\cite{wu2019wider} backbone.
    \item \textbf{MannequinChallenge}~\cite{li2019learning}.
    This dataset contains video clips captured using hand-held cameras, of people pretending frozen in a large variety of poses, imitating mannequins. We resize all frames to $1024\times 512$ and randomly split this dataset into 3040 train sequences and 292 test sequences, with sequence lengths ranging from 5-140 frames. We generate human body segmentation and part-specific UV coordinate maps using DensePose~\cite{Guler2018DensePose,wu2019detectron2} and body poses using OpenPose~\cite{cao2018openpose}.
    \item \textbf{ScanNet}~\cite{dai2017scannet}. This dataset contains multiple video clips captured in a total of 706 indoor rooms. We set aside 50 rooms for testing, and the rest for training. From each video sequence, we extracted 3 sub-sequences of length at most 100, resulting in 4000 train sequences and 289 test sequences, with images of size $512\times 512$. We used the provided segmentation maps based on the NYUDv2~\cite{silberman2012indoor} 40 labels.
\end{itemize}
For all datasets, we also use MegaDepth~\cite{li2018megadepth} to generate depth maps and add the visualized inverted depth images as input. As the MannequinChallenge and ScanNet datasets contain a large variety of objects and classes which are not fully annotated, we use edge maps produced by HED~\cite{xie2015holistically} in order to better represent the input content. In order to generate guidance images, we performed SfM on all the video sequences using OpenSfM~\cite{opensfm}, which provided 3D point clouds and estimated cameras poses and parameters as output.

\medskip
\noindent\textbf{Baselines.} We compare our method against the following strong baselines.
\begin{itemize}[label=\textbullet, topsep=2pt, itemsep=2pt]
    \item \vidtovid~\cite{wang2018video}. This is the prior \sota method for video-to-video synthesis. For comparison on Cityscapes, we use the publicly available pretrained model. For the other two datasets, we train \vidtovid from scratch using the public code, while providing the same input labels (semantic segmentation, depth, edge maps, \etc) as to our method.
    \item Inpainting~\cite{liu2018image}. We train a \sota partial convolution-based inpainting method to fill in the pixels missing from our guidance images.
    We train the models from scratch for each dataset, using masks obtained from the corresponding guidance images.
    \item \flow\,(World Consistency). As an ablation, we also compare against our model that does not use guidance images. In this case, only the first two SPADE layers in each Multi-SPADE block are trained (label and flow-warped previous output SPADEs). Other details are the same as our full model.
\end{itemize}

\input{tables/fid.tex}

\medskip
\noindent\textbf{Evaluation metrics.} We use both objective and subjective metrics for evaluating our model against the baselines.
\begin{itemize}[label=\textbullet, topsep=2pt, itemsep=2pt]
    \item \textit{Segmentation accuracy and Fr\'echet Inception Distance (FID)}. We adopt metrics widely used in prior work on image synthesis~\cite{chen2017photographic,park2019semantic,wang2018high} to measure the quality of generated video frames. We evaluate the output frames based on how well they can be segmented by a trained segmentation network. We report both the mean Intersection-Over-Union (mIOU) and Pixel Accuracy (P.A.) using the PSPNet~\cite{zhao2017pyramid} (Cityscapes) and DeepLabv2~\cite{chen2017deeplab} (MannequinChallenge \& ScanNet). We also use the Fr\'echet Inception Distance (FID)~\cite{heusel2017gans} to measure the distance between the distributions of the generated and real images, using the standard Inception-v3 network.
    \item \textit{Human preference score}. Using Amazon Mechanical Turk (AMT), we perform a subjective visual test to gauge the relative quality of videos. We evaluate videos on two criteria: 1) \textit{photorealism} and 2) \textit{temporal stability}. The first aims to find which generated video looks more like a real video, while the second aims to find which one is more temporally smooth and has lesser flickering. For each question, an AMT participant is shown two videos synthesized by two different methods, and asked to choose the better one according to the current criterion. We generate several hundred questions for each dataset, each of them is answered by 3 different workers. We evaluate an algorithm by the ratio that its outputs are preferred.
    \item \textit{Forward-Backward consistency}. A major contribution of our work is generating outputs that are consistent over a longer duration of time with the world that was previously generated. All our datasets have videos that explore new parts of the world over time, rarely revisiting previously explored parts. However, a simple way to revisit a location is to play the video in forward and then in reverse, \ie arrange frames from time $t=0, 1, \cdots, N-1, N, N-1, \cdots, 1, 0$. We can then compare the first produced and last produced frames and measure their difference. We measure the difference per-pixel in both RGB and LAB space, and a lower value would indicate better long-term consistency.
\end{itemize}

\input{figures/vid2vid_flow_guidance.tex}
\input{figures/fb_consistency.tex}

\medskip
\noindent\textbf{Main results.}
In Table~\ref{table:comparison}, we compare our proposed approach against vid2vid~\cite{wang2018video}, as well as SPADE~\cite{park2019semantic}, which is the single image generator that our method builds upon. We also compare against a version of our method that does not use guidance images and is thus not world-consistent (\flow). Inpainting~\cite{liu2018image} could not provide meaningful output images without large artifacts, as shown in Fig~\ref{fig:cityscapes_results}.
We can observe that our method consistently beats vid2vid on all three metrics on all three datasets, indicating superior image quality. Interestingly, our method also improves upon SPADE in FID, probably as a result of reducing temporal variance across an output video sequence. We also see improvements over \flow\ on almost all metrics.

In Table~\ref{table:human_pref}, we show human evaluation results on metrics of image realism and temporal stability. We observe that the majority of workers rank our method better on both metrics.

In Fig.~\ref{fig:cityscapes_results}, we visualize some sequences generated by the various methods (please zoom in and play the videos in Adobe Acrobat). We can observe that in the first row, vid2vid~\cite{wang2018video} produces temporal artifacts in the cars parked to the side and patterns on the road. SPADE~\cite{park2019semantic}, which produces one frame at a time, produces very unstable videos, as shown in the second row.
The third row shows outputs from the partial convolution-based inpainting~\cite{liu2018image} method. It clearly has a hard time producing visually and semantically meaningful outputs.
The fourth row shows \flow, an intermediate version of our method that uses labels and optical flow-warped previous output as input. While this clearly improves upon vid2vid in image quality and SPADE in temporal stability, it causes flickering in trees, cars, and signboards. The last row shows our method. Note how the textures of the cars, roads, and signboards, which are areas we have guidance images, are stable over time. We also provide high resolution, uncompressed videos for all three datasets on our website.

In Table~\ref{table:fb_consistency}, we compare the forward-backward consistency of different methods, and it shows that our method beats vid2vid~\cite{wang2018video} by a large margin, especially on the MannequinChallenge and ScanNet datasets (by more than a factor of 3). Figure~\ref{fig:fb_consistency} visualizes some frames at the start and end of generation. As can be seen, the outputs of vid2vid change dramatically, while ours are consistent. We show additional qualitative examples in Fig.~\ref{fig:more_results}. We also provide additional quantitative results on short-term consistency in Appendix~\ref{appendix:results}.

\input{figures/more_fb_results.tex}

%% file: tables/fid.tex

\begin{table}[t]
    \setlength{\tabcolsep}{4pt}
    \centering
    \caption{Comparison scores. $\downarrow$ means lower is better, while $\uparrow$ means the opposite.}
    \label{table:comparison}
    \resizebox{\columnwidth}{!}{%
        \begin{tabular}{l|ccc|ccc|ccc}
            \hline
            \multirow{2}{*}{Method} & \multicolumn{3}{c|}{Cityscapes} & \multicolumn{3}{c|}{MannequinChallenge} & \multicolumn{3}{c}{ScanNet} \\
            & FID$\downarrow$ & mIOU$\uparrow$ & P.A.$\uparrow$ & FID$\downarrow$ & mIOU$\uparrow$ & P.A.$\uparrow$ & FID$\downarrow$ & mIOU$\uparrow$ & P.A.$\uparrow$ \\\hline
            \multicolumn{10}{c}{\bf Image synthesis models} \\
            SPADE~\cite{park2019semantic} & 48.25 & 0.63 & 0.95  & 29.99 & 0.13 & 0.63 & 31.46 & 0.08 & 0.54 \\ \hline
            \multicolumn{10}{c}{\bf Video synthesis models} \\
            \vidtovid~\cite{wang2018video} & 69.07 & 0.55 & 0.94 & 72.25 & 0.05 & 0.45 & 60.03 & 0.04 & 0.35 \\
            \flow & 51.51 & {\bf 0.62} & 0.95 & 27.23 & 0.17 & 0.67 & {\bf 20.93} & 0.12 & 0.62 \\
            \guidance & {\bf 49.89} & 0.61 & {\bf 0.95} & {\bf 22.69} & {\bf 0.19} & {\bf 0.69} & 21.07 & {\bf 0.13} & {\bf 0.63} \\
            \hline
        \end{tabular}
    }

    \setlength{\tabcolsep}{4pt}
    \centering
    \caption{Human preference scores. Higher is better.}
    \label{table:human_pref}
    \resizebox{0.75\columnwidth}{!}{%
    \begin{tabular}{lccc}
        \hline
        Compared Methods & Cityscapes & MannequinChallenge & ScanNet \\\hline
        \multicolumn{4}{c}{\bf Image Realism} \\
        \guidance/\vidtovid~\cite{wang2018video} & {\bf 0.73}/0.27 & {\bf 0.83}/0.17 & {\bf 0.77}/0.23 \\
        \hline
        \multicolumn{4}{c}{\bf Temporal Stability} \\
        \guidance/\vidtovid~\cite{wang2018video} & {\bf 0.75}/0.25 & {\bf 0.63}/0.37 & {\bf 0.82}/0.18 \\\hline
    \end{tabular}
    }

    \setlength{\tabcolsep}{4pt}
	\centering
	\caption{Forward-backward consistency. $\triangle$ means difference.}
	\label{table:fb_consistency}
    \resizebox{\columnwidth}{!}{%
	\begin{tabularx}{\textwidth}{ l| YY|YY|YY }
		\hline
		\multirow{2}{*}{Method} & \multicolumn{2}{c|}{Cityscapes} & \multicolumn{2}{c|}{MannequinChallenge} & \multicolumn{2}{c}{ScanNet} \\
		& $\triangle$RGB$\downarrow$ &  $\triangle$LAB$\downarrow$ &  $\triangle$RGB$\downarrow$ & $\triangle$LAB$\downarrow$  &  $\triangle$RGB$\downarrow$ & $\triangle$LAB$\downarrow$ \\\hline
		\vidtovid~\cite{wang2018video} & 14.90 & 3.46 & 37.56 & 9.42 & 46.30 & 12.16 \\
		\guidance & {\bf 8.73} & {\bf 2.04} & {\bf 12.61} & {\bf 3.61} & {\bf 11.85} & {\bf 3.41} \\\hline
	\end{tabularx}
    }
\end{table}

%% file: figures/vid2vid_flow_guidance.tex

\begin{table}[t]
    \setlength{\tabcolsep}{0.25pt}
    \centering

    \resizebox{0.9\columnwidth}{!}{%
    \begin{tabular}{ccc}
        {\small \rotatebox[origin=c]{\rotangle}{vid2vid~\cite{wang2018video}}} &
        \raisebox{-.5\height}{\href{\website stuttgart_02_06_vid2vid.mp4}{\includegraphics[width=0.49\textwidth, trim= 0 3cm 0 0]{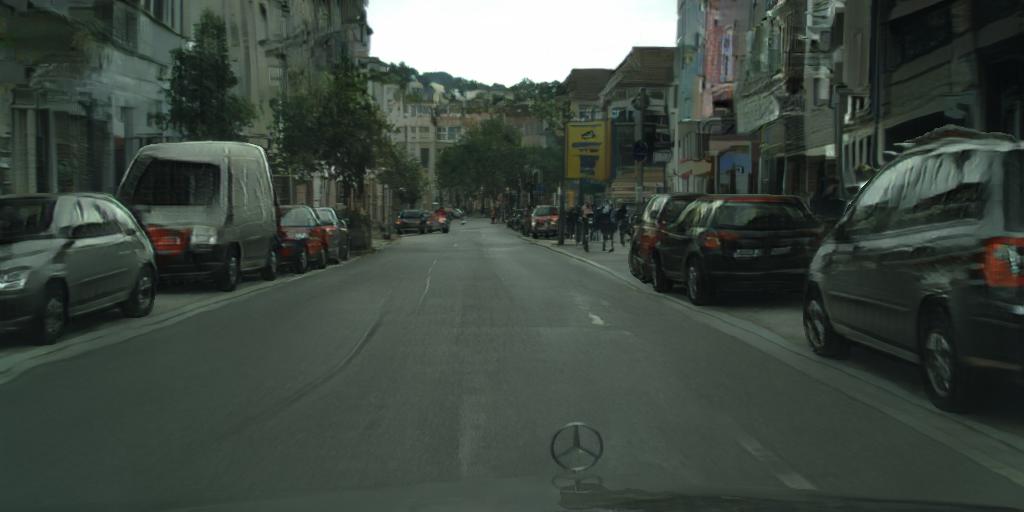}}} &
        \raisebox{-.5\height}{\href{\website stuttgart_01_06_vid2vid.mp4}{\includegraphics[width=0.49\textwidth, trim= 0 3cm 0 0]{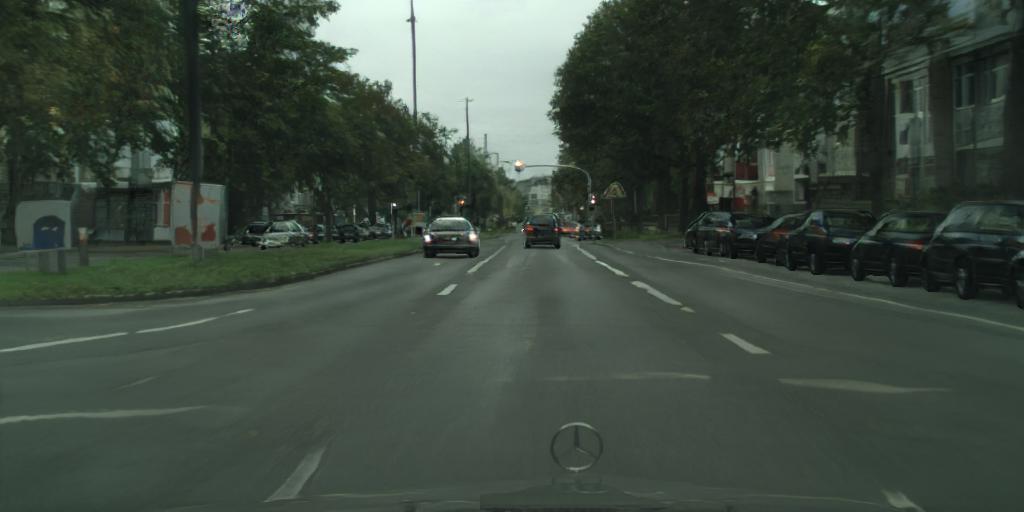}}} \\

        {\small \rotatebox[origin=c]{\rotangle}{SPADE~\cite{park2019semantic}}} &
        \raisebox{-.5\height}{\href{\website stuttgart_02_06_single.mp4}{\includegraphics[width=0.49\textwidth, trim= 0 3cm 0 0]{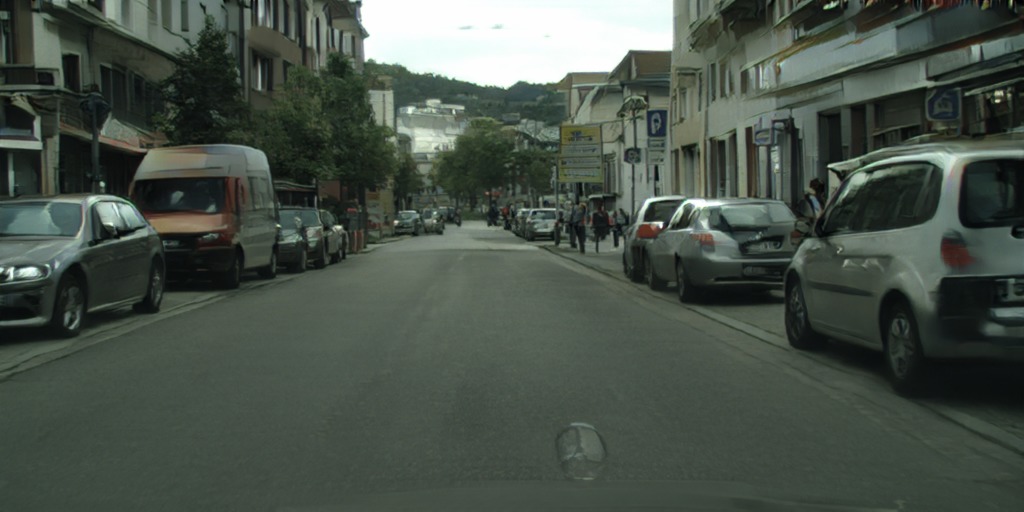}}} &
        \raisebox{-.5\height}{\href{\website stuttgart_01_06_single.mp4}{\includegraphics[width=0.49\textwidth, trim= 0 3cm 0 0]{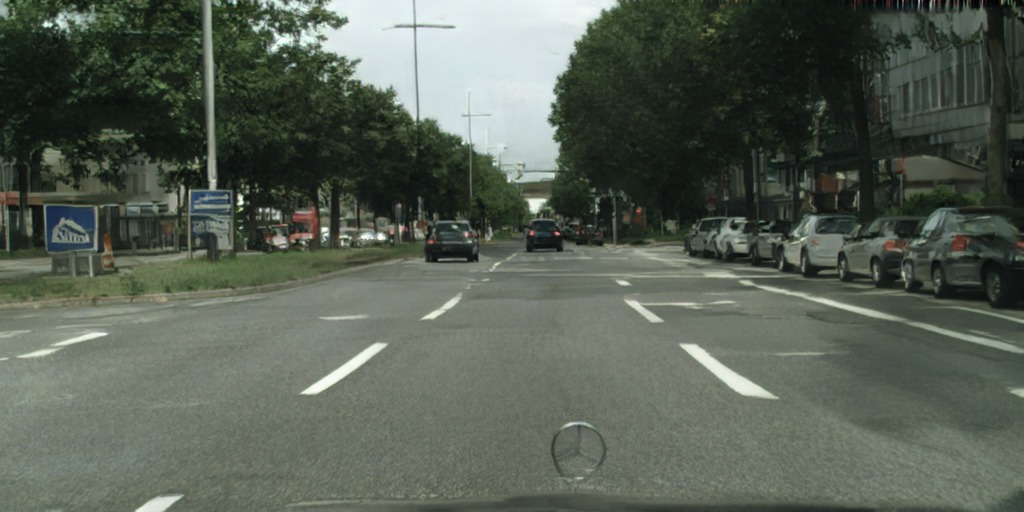}}} \\

        {\small \rotatebox[origin=c]{\rotangle}{Inpainting~\cite{liu2018image}}} &
        \raisebox{-.5\height}{\href{\website stuttgart_02_06_inpainting.mp4}{\includegraphics[width=0.49\textwidth, trim= 0 3cm 0 0]{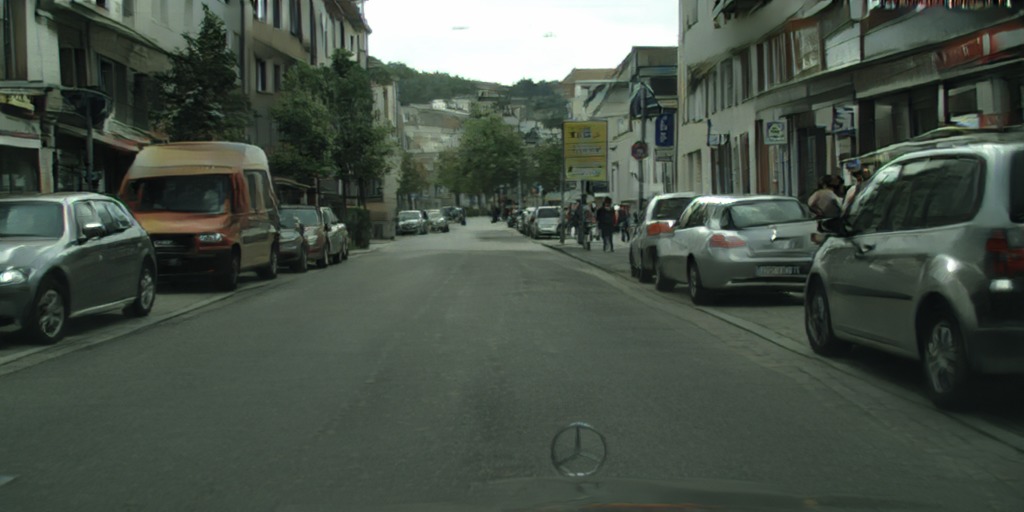}}} &
        \raisebox{-.5\height}{\href{\website stuttgart_01_06_inpainting.mp4}{\includegraphics[width=0.49\textwidth, trim= 0 3cm 0 0]{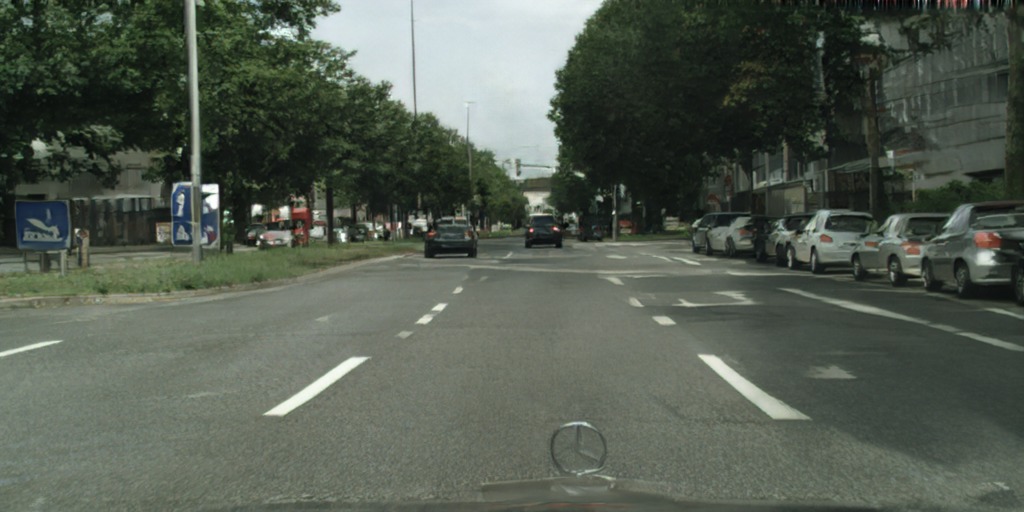}}} \\

        {\small \rotatebox[origin=c]{\rotangle}{\flow}} &
        \raisebox{-.5\height}{\href{\website stuttgart_02_06_flow.mp4}{\includegraphics[width=0.49\textwidth, trim= 0 3cm 0 0]{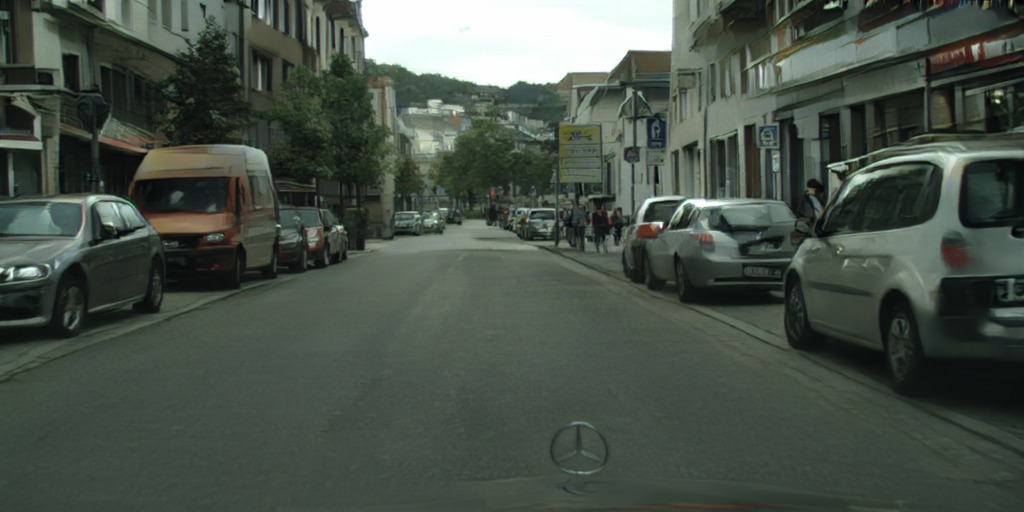}}} &
        \raisebox{-.5\height}{\href{\website stuttgart_01_06_flow.mp4}{\includegraphics[width=0.49\textwidth, trim= 0 3cm 0 0]{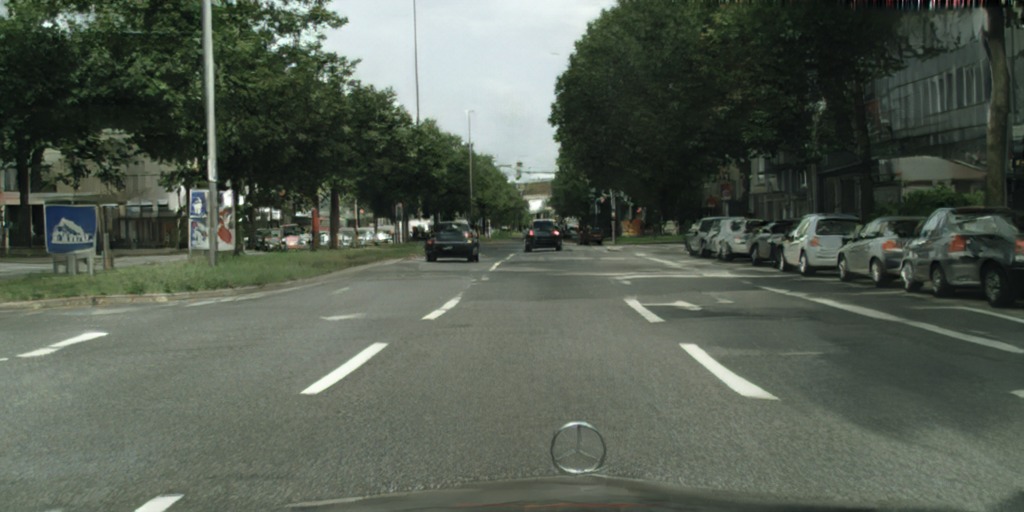}}} \\

        {\small \rotatebox[origin=c]{\rotangle}{\guidance}} &
        \raisebox{-.5\height}{\href{\website stuttgart_02_06_guidance.mp4}{\includegraphics[width=0.49\textwidth, trim= 0 3cm 0 0, clip]{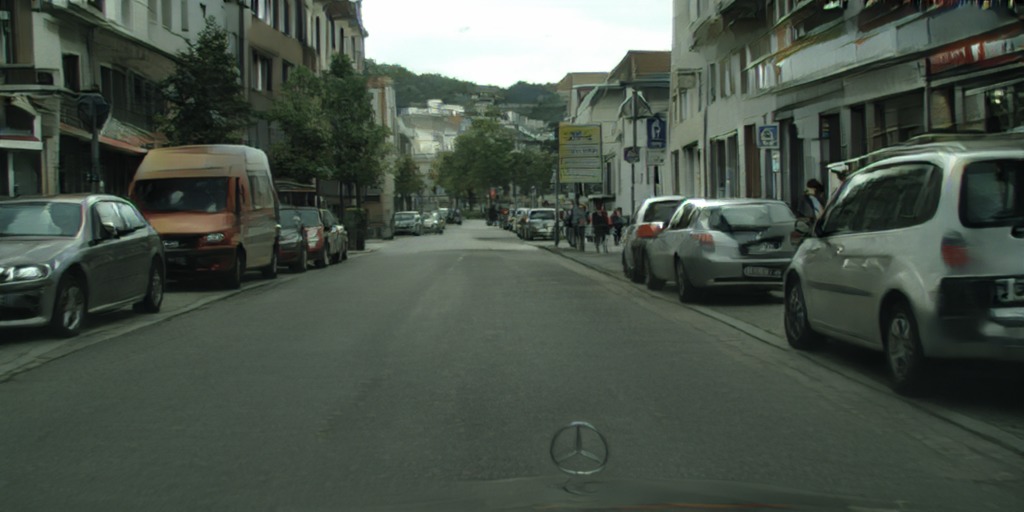}}} &
        \raisebox{-.5\height}{\href{\website stuttgart_01_06_guidance.mp4}{\includegraphics[width=0.49\textwidth, trim= 0 3cm 0 0, clip]{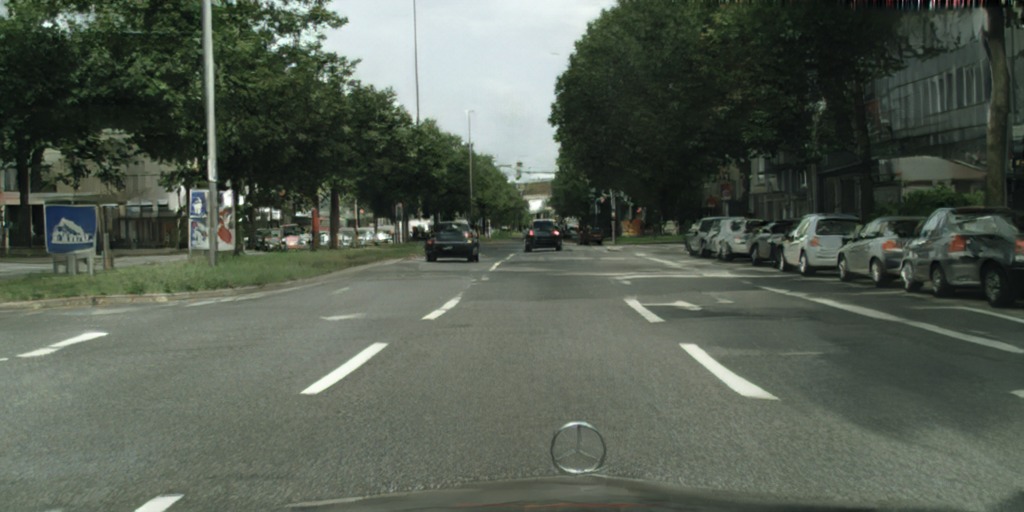}}} \\
    \end{tabular}
    }

    \captionof{figure}{Comparison of different video generation methods on the Cityscapes dataset. Note that for our results, the textures of the cars, roads, and signboards are stable over time, while they change gradually in vid2vid and other methods.
    Click on an image to play the video.}
    \label{fig:cityscapes_results}
\end{table}

%% file: figures/fb_consistency.tex

\begin{table}[t]
    \setlength{\tabcolsep}{0.2pt}
    \centering

    \resizebox{\columnwidth}{!}{%
    \begin{tabular}{cccc}
        & Cityscapes & MannequinChallenge & ScanNet \\

        {\small \rotatebox[origin=c]{\rotangle}{\vidtovid~\cite{wang2018video}}} &
        \raisebox{-.5\height}{\href{\website fb_pairs/cityscapes_vid2vid.gif}{\includegraphics[height=2cm, trim= 0 3cm 0 0, clip]{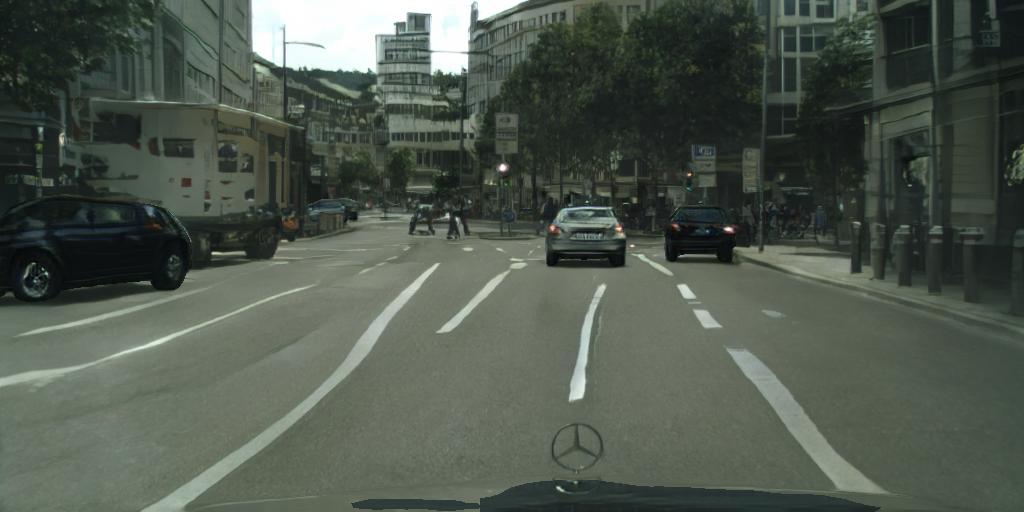}}} &
        \raisebox{-.5\height}{\href{\website fb_pairs/mannequin_vid2vid.gif}{\includegraphics[height=2cm]{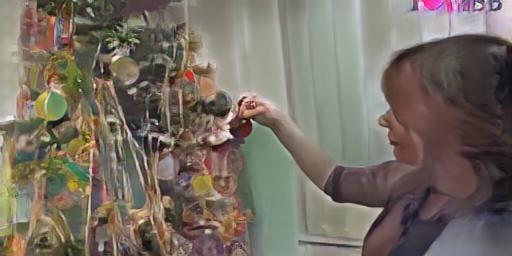}}} &
        \raisebox{-.5\height}{\href{\website fb_pairs/scannet_vid2vid.gif}{\includegraphics[height=2cm]{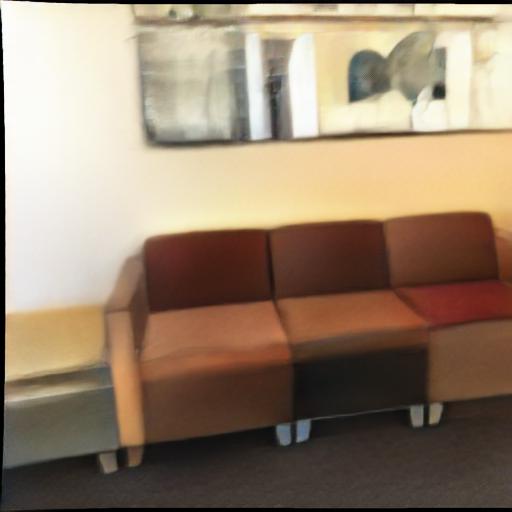}}} \\

        {\small \rotatebox[origin=c]{\rotangle}{\guidance}} &
        \raisebox{-.5\height}{\href{\website fb_pairs/cityscapes_guidance.gif}{\includegraphics[height=2cm, trim= 0 3cm 0 0, clip]{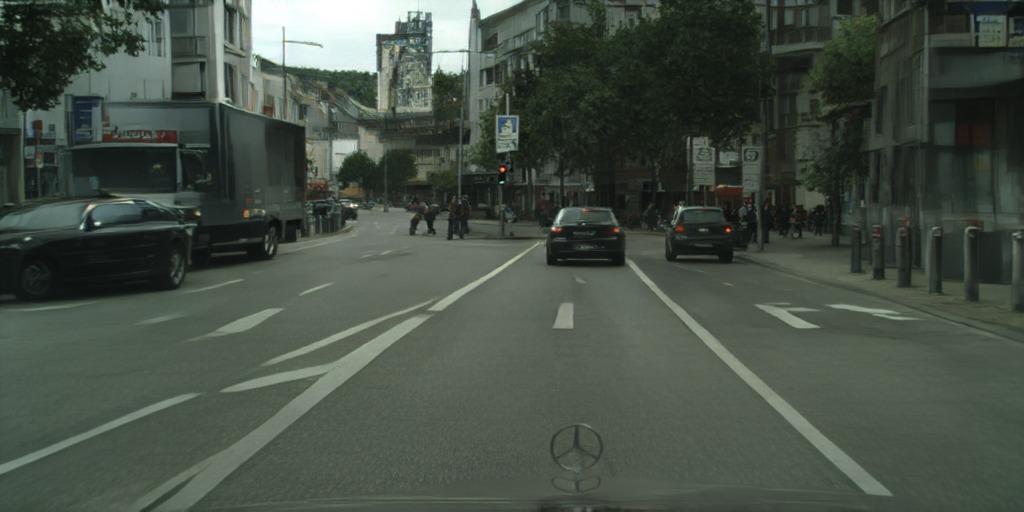}}} &
        \raisebox{-.5\height}{\href{\website fb_pairs/mannequin_guidance.gif}{\includegraphics[height=2cm]{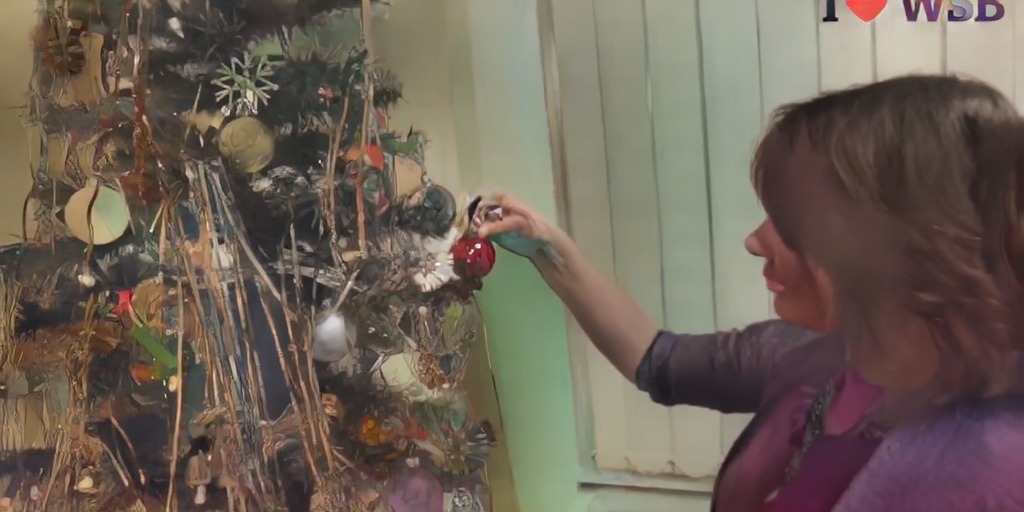}}} &
        \raisebox{-.5\height}{\href{\website fb_pairs/scannet_guidance.gif}{\includegraphics[height=2cm]{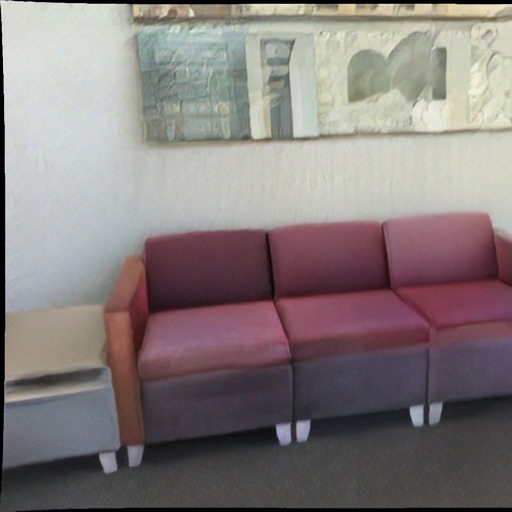}}}

    \end{tabular}
    }
    \captionof{figure}{Forward-backward consistency. Click on each image to see the change in output when the viewpoint is revisited. Note how drastically vid2vid results change, while ours remain almost the same.}
    \label{fig:fb_consistency}
\end{table}

%% file: figures/more_fb_results.tex

\begin{table}[t]
    \setlength{\tabcolsep}{0.2pt}
    \centering

    \resizebox{\columnwidth}{!}{%
    \begin{tabular}{cccc}
        \href{\website mann_scan/01_n.mp4}{\includegraphics[height=1.5cm, trim={2cm 0 3cm 0}]{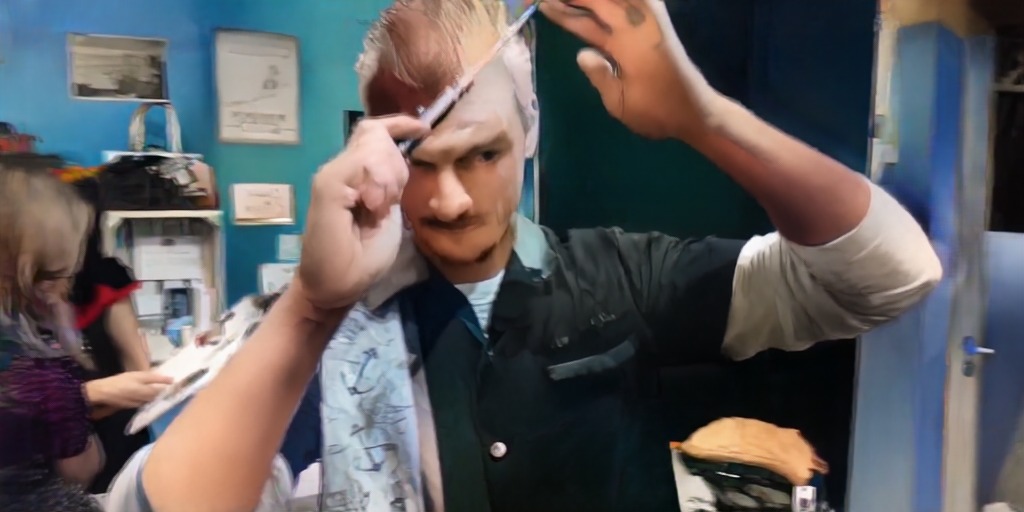}} &
        \href{\website mann_scan/02_n.mp4}{\includegraphics[height=1.5cm, trim={4cm 0 3cm 0}]{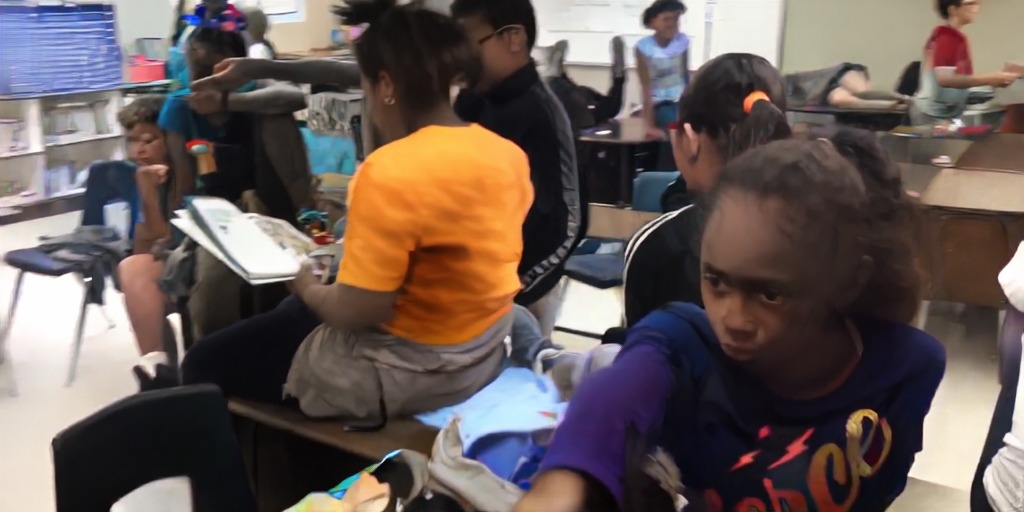}} &
        \href{\website mann_scan/03_n.mp4}{\includegraphics[height=1.5cm]{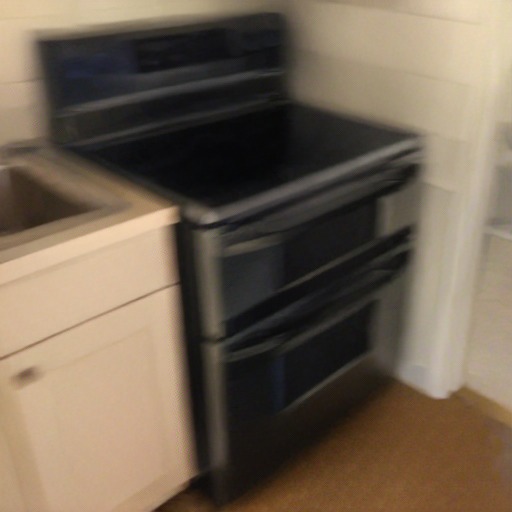}} &
        \href{\website mann_scan/04_n.mp4}{\includegraphics[height=1.5cm]{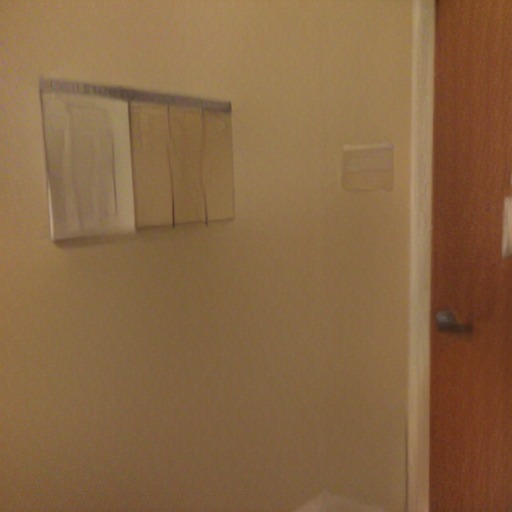}}

    \end{tabular}
    }
    \captionof{figure}{Qualitative results on the MannequinChallenge and ScanNet datasets. Click on an image to play video. Note the results are consistent over time and viewpoints.}
    \label{fig:more_results}
\end{table}

%% file: src/applications.tex

\input{figures/stereo.tex}

\medskip
\noindent\textbf{Generating consistent stereo outputs.} Here, we show a novel application enabled by our method through the use of guidance images. We show videos rendered simultaneously for multiple viewpoints, specifically for a pair of stereo viewpoints on the Cityscapes dataset in Fig.~\ref{fig:stereo_outputs}. For the strongest baseline, \flow, the left-right videos can only be generated independently, and they clearly are not consistent across multiple viewpoints, as highlighted by the boxes. On the other hand, our method can generate left-right videos in sync by sharing the underlying 3D point cloud and guidance maps. Note how the textures on roads, including shadows, move in sync and remain consistent over time and camera locations.

%% file: figures/stereo.tex

\begin{table}[t]
    \setlength{\tabcolsep}{0.5pt}
    \centering

    \resizebox{\columnwidth}{!}{%
    \begin{tabular}{cc}
        {\small \rotatebox[origin=c]{\rotangle}{\flow}} &
        \raisebox{-.5\height}{\href{\website stereo/stereo_stuttgart_00_346_flow.mp4}{\includegraphics[width=\textwidth, trim= 0 3cm 0 0, clip]{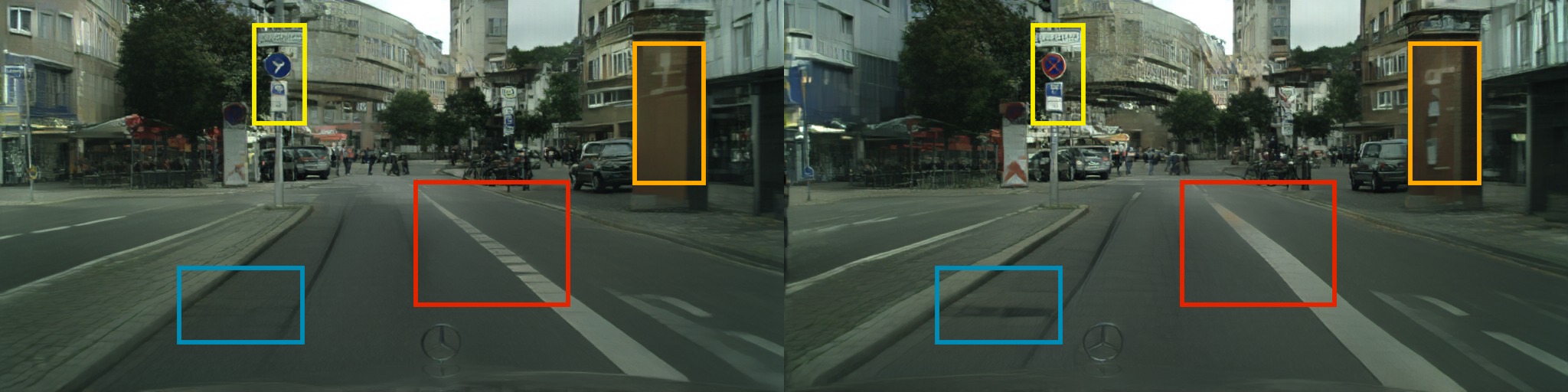}}} \\

        & \\[-10pt]

        {\small \rotatebox[origin=c]{\rotangle}{\guidance}} &
        \raisebox{-.5\height}{\href{\website stereo/stereo_stuttgart_00_346_guidance.mp4}{\includegraphics[width=\textwidth, trim= 0 3cm 0 0, clip]{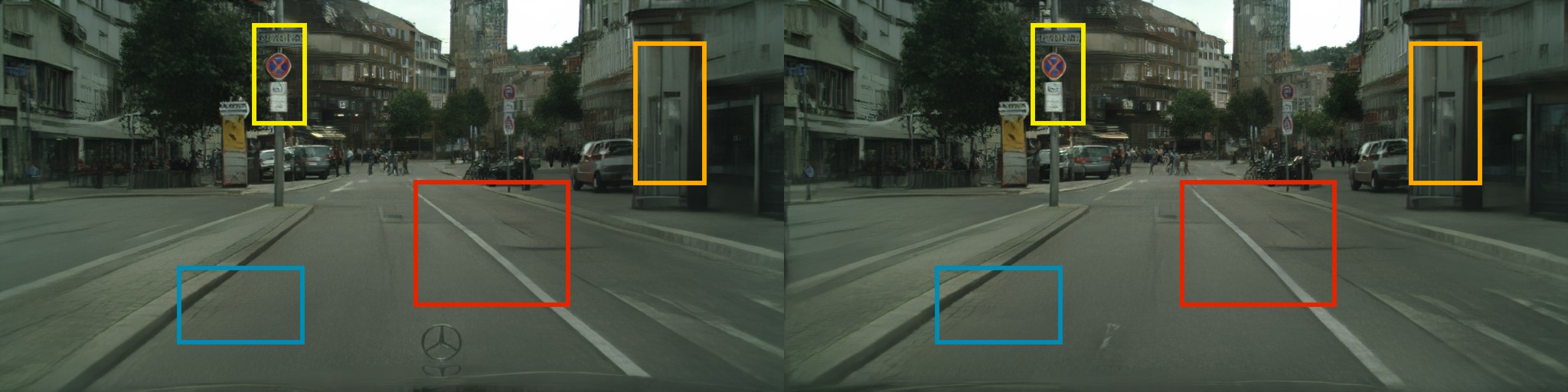}}}

    \end{tabular}
    }

    \captionof{figure}{Stereo results on Cityscapes. Click on an image to see the the outputs produced by a pair of stereo cameras. Note how our method produces images consistent across the two views, while they differ in the highlighted regions without using the world consistency.}
    \label{fig:stereo_outputs}
\end{table}

%% file: src/conclusions.tex

\section{Conclusions and discussion}
We presented a video-to-video synthesis framework that can achieve world consistency. By using a novel guidance image extracted from the generated 3D world, we are able to synthesize the current frame conditioned on all the past frames. The conditioning was implemented using a novel Multi-SPADE module, which not only led to better visual quality, but also made transplanting a single image generator to a video generator possible. Comparisons on several challenging datasets showed that our method improves upon prior \sota methods.

While advancing the state-of-the-art, our framework still has several limitations. For example, the guidance image generation is based on SfM. When SfM fails to register the 3D content, our method will also fail to ensure consistency. Also, we do not consider a possible change in time of the day or lighting in the current framework.
In the future, our framework can benefit from improved guidance images enabled by better 3D registration algorithms. Furthermore, the albedo and shading of the 3D world may be disentangled to better model the time effects. We leave these to future work.

\medskip\noindent{\bf Acknowledgements.} We would like to thank Jan Kautz, Guilin Liu, Andrew Tao, and Bryan Catanzaro for their feedback, and Sabu Nadarajan, Nithya Natesan, and Sivakumar Arayandi Thottakara for helping us with the compute, without which this work would not have been possible.

%% file: objectives.tex
\section{Objective functions}
\label{appendix:objective_functions}

Our objective functions contain five losses: an image GAN loss, a video GAN loss, a perceptual loss, a flow-warping loss, and a world-consistency loss. Except for the world-consistency loss, the others are inherited from the \vidtovid~\cite{wang2018video}. Note that we replace the least square losses used in the \vidtovid for GAN losses  with the hinge losses as used in SPADE~\cite{park2019semantic}. We describe these terms in details in the following.

\medskip
\noindent{\bf GAN losses.} Let $\bs_1^T \equiv \{ \bs_{1},\bs_{2},...,\bs_{T}\}$ be a sequence of input semantic frames. Let $\bx_1^T \equiv \{\bx_{1},\bx_{2},...,\bx_{T}\}$ be the sequence of corresponding real video frames, and $\x_1^T \equiv \{\x_{1},\x_{2},...,\x_{T}\}$ be the synthesized frames by our generator. Define $(\bx_t,\bs_t)$ as one pair of frames at a particular time instance where $\bx_t\in\bx_1^T$ and $\bs_t\in\bs_1^T$. The image GAN loss ($\mathcal{L}_{I}^t$) and the video GAN loss ($\mathcal{L}_{V}^t$) for time $t$ are then defined as
\begin{align}
\mathcal{L}_{I}^t =& E_{(\bx_t,\bs_t)}[\min(0,-1+D_I(\bx_t,\bs_t))] + \\ &E_{(\x_t,\bs_t)}[\min(0,-1-D_I(\x_t,\bs_t)] \\
\mathcal{L}_{V}^t =& E_{\bx_{t-K+1}^{t}}[\min(0,-1+D_V(\bx_{t-K+1}^{t})] + \\
&E_{\x_{t-K}^{t-1}}[\min(0,-1-D_V(\x_{t-K}^{t-1}))]
\end{align}
where $D_I$ and $D_V$ are the image and video discriminators, respectively. The video discriminator takes $K$ consecutive frames and concatenates them together for discrimination.
For both GAN losses, we also accompany them by the feature matching loss ($\mathcal{L}_{FM}^t$) as in pix2pixHD~\cite{wang2018high},
\begin{equation}
    \mathcal{L}_{FM,I/V}^t =  \sum_{i}\frac{1}{P_i}\left[ ||D_{\{I/V\}}^{(i)}(\bx_t)-D_{\{I/V\}}^{(i)}(\x_t)||_1 \right],
\end{equation}
where $D_{\{I/V\}}^{(i)}$ denotes the $i$-th layer with $P_i$ elements of the discriminator network $D_I$ or $D_V$.

\input{figuretex/embedding}

\medskip
\noindent{\bf Perceptual loss.} We use the VGG-16 network~\cite{simonyan2014very} as a feature extractor and minimize L1 losses between the extracted features from the real and the generated images. In particular,
\begin{equation}
    \mathcal{L}_{P}^t = \sum_{i}\frac{1}{P_i}\left[ ||\psi^{(i)}(\bx_t)-\psi^{(i)}(\x_t)||_1 \right],
\end{equation}
where $\psi^{(i)}$ denotes the $i$-th layer of the VGG network.

\medskip
\noindent{\bf Flow-warping loss.} We first warp the previous frame to the current frame using optical flow. We then encourage the warped frame to be similar to the current frame by using an L1 loss,
\begin{equation}
    \mathcal{L}_{F}^t = ||\x_t-\bw_t(\x_{t-1})||_1
\end{equation}
where $\bw_t$ is the warping function derived from optical flow.

\medskip
\noindent{\bf World-consistency loss.} Finally, we add the world consistency by enforcing the generated image to be similar to our guidance image. It is achieved by
\begin{equation}
    \mathcal{L}_{WC}^t = ||\x_t-\g_t||_1
\end{equation}
where $\g_t$ is our estimated guidance image.\\

\noindent The overall objective function is then
\begin{align}
    \mathcal{L} = \displaystyle\sum_t & \min_{G} \left( \max_{D_I,D_V} (\lambda_{I}\mathcal{L}_{I}^t + \lambda_{V}\mathcal{L}_{V}^t) \right) +  \\
    & \min_{G} \left(\lambda_{FM}\mathcal{L}_{FM}^t + \lambda_{P}\mathcal{L}_{P}^t + \lambda_{F}\mathcal{L}_{F}^t + \lambda_{W}\mathcal{L}_{WC}^t \right)
\end{align}
where $\lambda$ are the weights for each individual terms, which are set to 1, 1, 10, 10, 10, 10 in all of our experiments.

\medskip
\noindent{\bf Optimization details.} We use the ADAM optimizer~\cite{kingma2014adam} with $(\beta_1, \beta_2) = 0, 0.999$ for all experiments and network components.
We use a learning rate of 1e-4 for the encoder and generator networks (which are described below) and 4e-4 for the discriminators.

%% file: figuretex/embedding.tex
\begin{figure}[!t]
    \centering
    \includegraphics[width=0.4\textwidth]{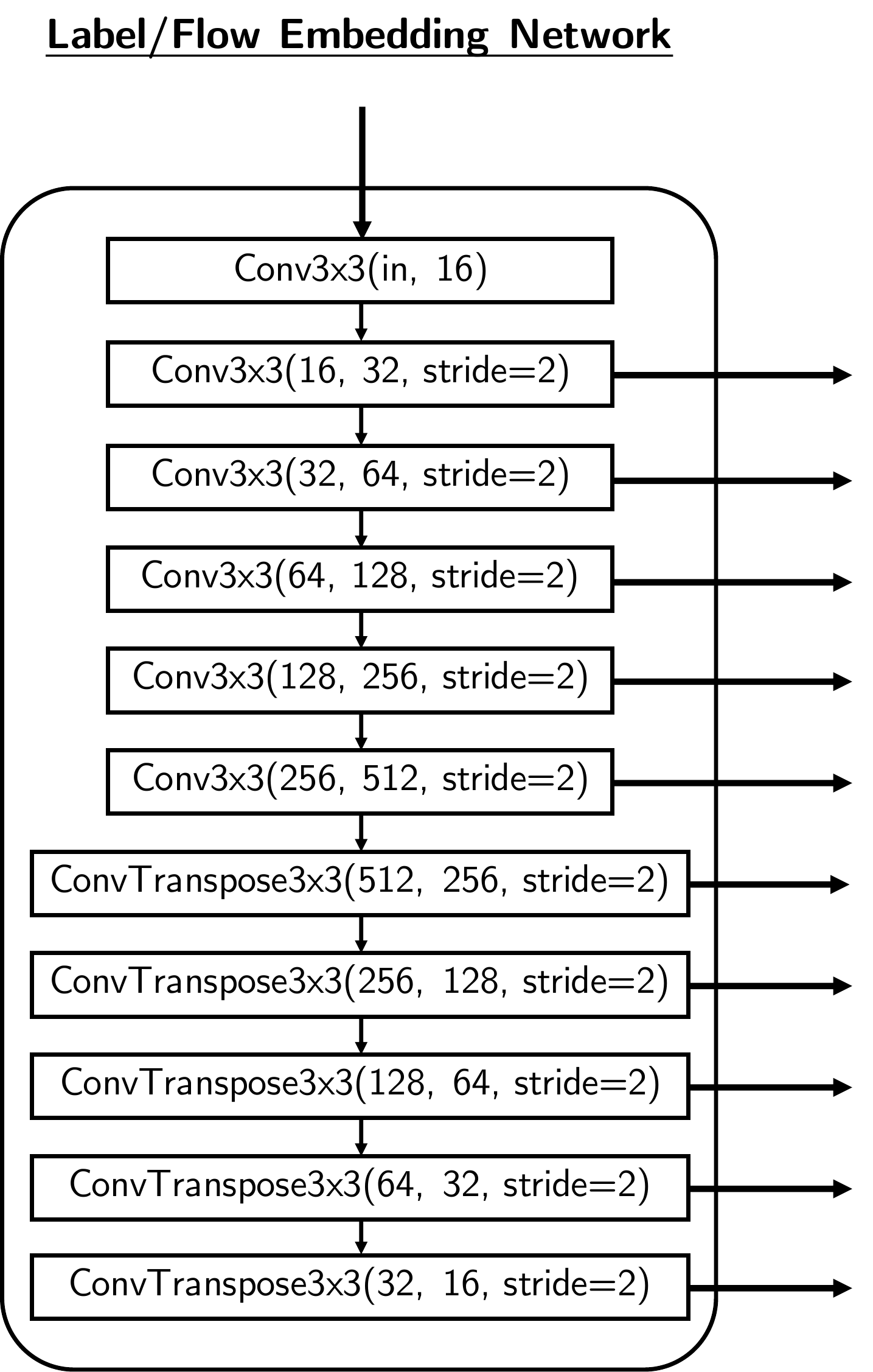}
    \caption{Label / flow-warped image embedding network. }
    \label{fig:embedding}
\end{figure}

%% file: network.tex
\section{Network architecture}
\label{appendix:network}

\input{figuretex/encoder}

As described in the main paper, our framework contains four components: a label embedding network (Fig.~\ref{fig:embedding}), an image encoder (Fig.~\ref{fig:encoder}), a flow embedding network (Fig.~\ref{fig:embedding}), and an image generator (Fig.~\ref{fig:generator}).

\medskip
\noindent{\bf Label embedding network (Fig.~\ref{fig:embedding}).} We adopt an encoder-decoder style network to embed the input labels into different feature representations, which are then fed to the Multi-SPADE modules in the image generator.

\medskip
\noindent{\bf Image / segmentation encoder (Fig.~\ref{fig:encoder}).} These networks generate the input to the main image generator. The segmentation encoder is used when generator the first frame in the sequence, while the image encoder is used when generating the subsequent frames. The segmentation encoder encodes the input semantics of the first frame, while the image encoder encodes the previously generated frame.

\input{figuretex/generator}

\input{figuretex/multispade}
\medskip
\noindent{\bf Flow embedding network (Fig.~\ref{fig:embedding}).} It is used to embed the optical flow-warped previous frame, which adopts the same architecture as the label embedding network except for the number of input channels. The embedded features are again fed to the Multi-SPADE layers in the main image generator.

\medskip
\noindent{\bf Image generator (Fig.~\ref{fig:generator}).} The generator consists of a series of Multi-SPADE residual blocks (M-SPADE ResBlks) and upsampling layers. The structure of each M-SPADE Resblk is shown in Fig.~\ref{fig:mspade_resblk}, which replaces the SPADE layers in the original SPADE Resblks with Multi-SPADE layers.

\medskip
\noindent{\bf Discriminators.} We use the same image and video discriminators as vid2vid~\cite{wang2018video}.

%% file: figuretex/encoder.tex
\begin{figure}[!t]
    \centering
    \includegraphics[width=0.7\textwidth]{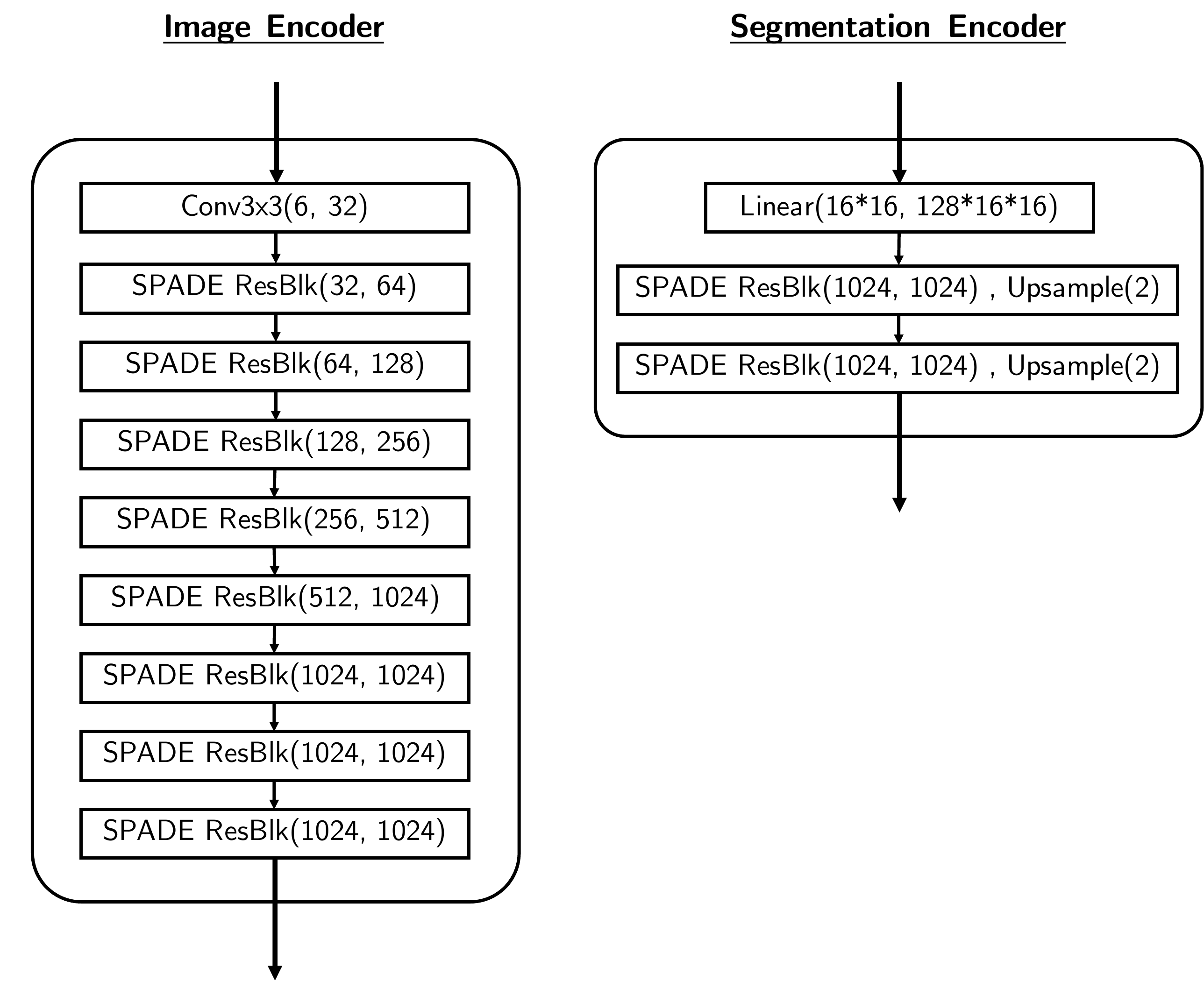}
    \caption{Previous image / segmentation encoder.}
    \label{fig:encoder}
\end{figure}

%% file: figuretex/generator.tex
\begin{figure}[!t]
    \centering
    \includegraphics[width=0.6\textwidth]{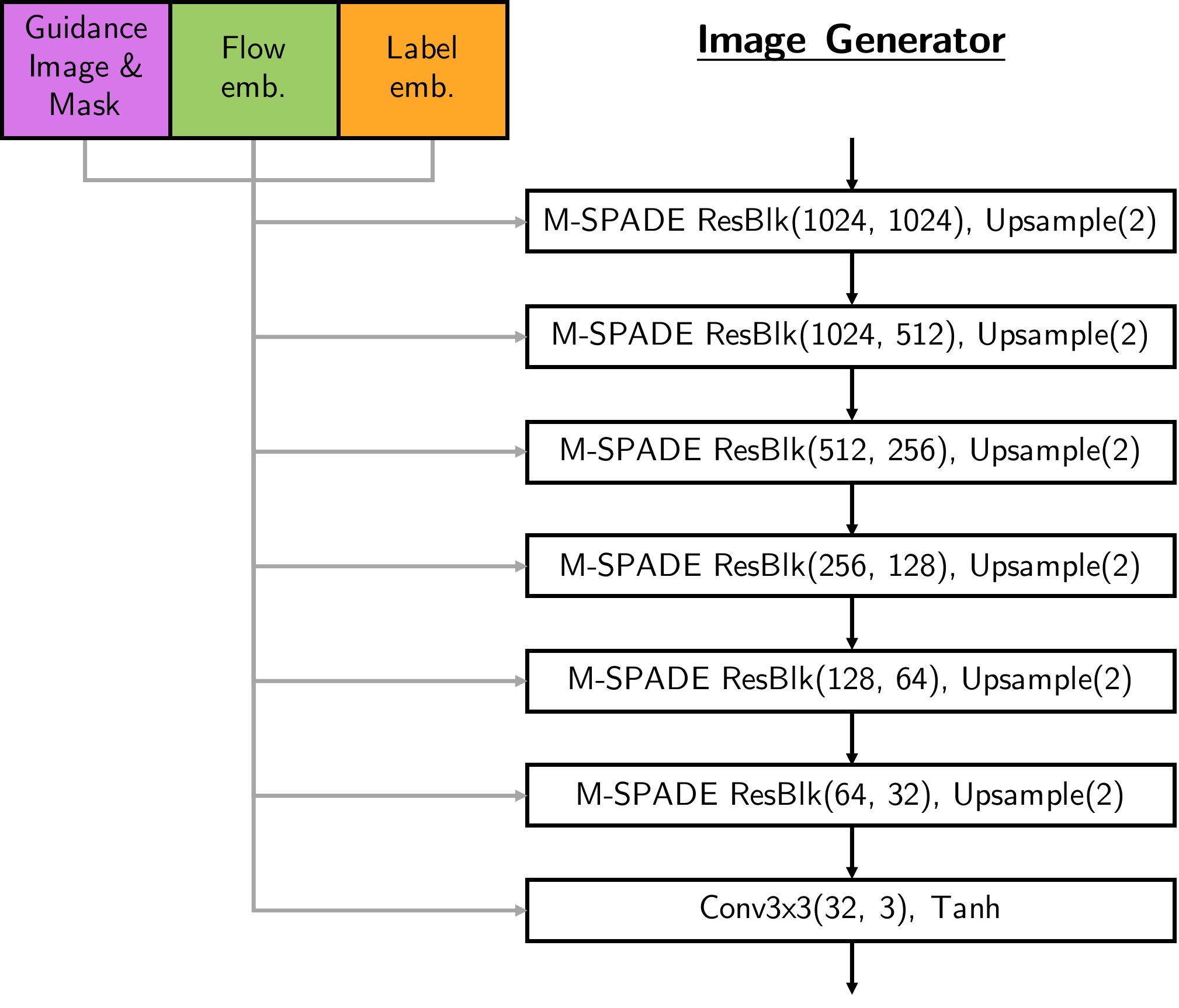}
    \caption{Image generator.}
    \label{fig:generator}
\end{figure}



%% file: figuretex/multispade.tex
\begin{figure}[!t]
    \centering
    \includegraphics[width=\textwidth]{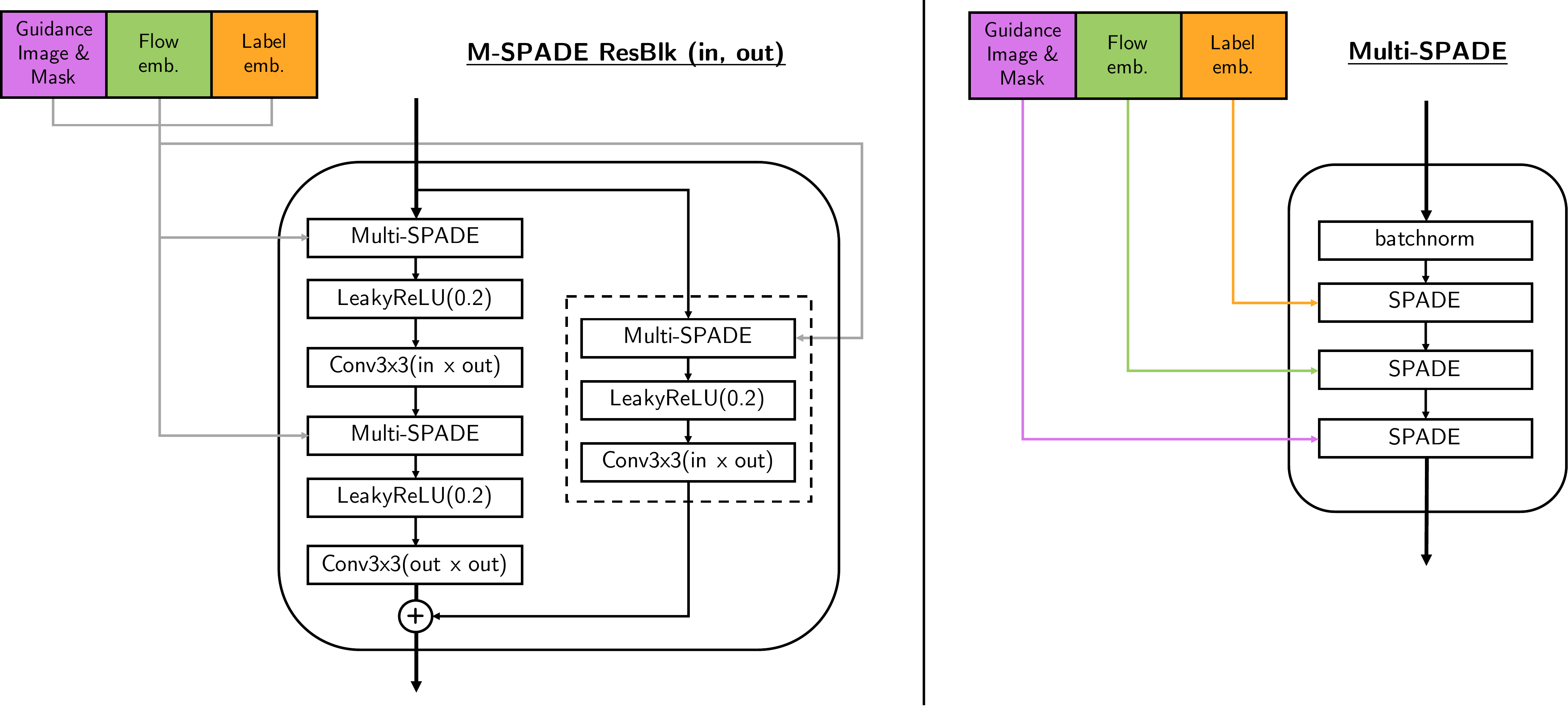}
    \caption{Multi-SPADE Residual Block and Multi-SPADE module.}
    \label{fig:mspade_resblk}
\end{figure}

%% file: additional_results.tex

\section{Additional Results}
\label{appendix:results}

\noindent{\bf Short-term temporal video consistency.}
For each sequence, we first take two neighboring frames from the ground truth images to compute the optical flow between them using FlowNet2~\cite{ilg2017flownet}. We then use the optical flow to warp the corresponding synthesized images and compute the L1 distance between the warped image and the target image, in RGB space, normalized by the number of pixels and channels. This process is repeated for all pairs of neighboring frames in all sequences and averaged. The result is shown in below in Table~\ref{table:short_term_consistency}. As can be seen, Ours w/o World Consistency (W.C.) consistently performs better than vid2vid~\cite{wang2018video}, and Ours (with world consistency) again consistently outperforms Ours w/o W.C.

\begin{table}[h!]
    \centering
    \caption{Short-term temporal consistency scores. Lower is better.}
    \label{table:short_term_consistency}
    \begin{tabular}{l|c|c|c}
        \hline
        Dataset & vid2vid~\cite{wang2018video} & Ours w/o W.C. & Ours \\\hline
        Cityscapes & 0.0036 & 0.0032 & {\bf 0.0029} \\
        MannequinChallenge & 0.0397 & 0.0319 & {\bf 0.0312} \\
        ScanNet & 0.0351 & 0.0278 & {\bf 0.0192} \\\hline
    \end{tabular}
\end{table}